\documentclass{article}

\PassOptionsToPackage{numbers, compress}{natbib}

\usepackage[final]{neurips_2025}

\usepackage[utf8]{inputenc} 
\usepackage[T1]{fontenc}    
\usepackage{url}            
\usepackage{booktabs}       
\usepackage{amsfonts}       
\usepackage{nicefrac}       
\usepackage{microtype}      
\usepackage{bm}
\usepackage{mathtools}
\usepackage[table]{xcolor}
\usepackage{multirow}
\usepackage{wrapfig}  
\usepackage{wrapfig}    
\usepackage{lineno}   
\usepackage{array}    
\usepackage[hidelinks,breaklinks=true,colorlinks,citecolor=citecolor,linkcolor=linkcolor]{hyperref}
\definecolor{citecolor}{HTML}{2980b9}
\definecolor{linkcolor}{HTML}{c0392b}
\definecolor{sem}{HTML}{2E75B6}
\definecolor{tok}{HTML}{F3B000}

\definecolor{cgreen}{HTML}{70AD47}
\definecolor{cblue}{HTML}{4D96FF}

\title{\textit{Video-RAG}: Visually-aligned Retrieval-Augmented Long Video Comprehension}

\author{%
  Yongdong Luo$^{1}$ \quad Xiawu Zheng$^{1}$\thanks{Corresponding author: zhengxiawu@xmu.edu.cn} \quad Guilin Li$^{1}$ \quad Shukang Yin \quad Haojia Lin$^{1}$ \\  \textbf{Chaoyou Fu}$^2$ \quad \textbf{Jinfa Huang}$^3$ \quad \textbf{Jiayi Ji}$^1$ \quad
   \textbf{Fei Chao}$^1$ \quad \textbf{Jiebo Luo}$^3$ \quad \textbf{Rongrong Ji}$^{1}$
  \\[0.2cm] 
  $^1$Key Laboratory of Multimedia Trusted Perception and Efficient Computing, \\ \; Ministry of Education of China, Xiamen University, 361005, P.R. China \; \\
  $^2$ Nanjing University \;
  $^3$ University of Rochester\\
}

\begin{document}

\maketitle

\begin{abstract}
\label{sec:abs}

Existing large video-language models (LVLMs) struggle to comprehend long videos correctly due to limited context. 
To address this problem, fine-tuning long-context LVLMs and employing GPT-based agents have emerged as promising solutions. 
However, fine-tuning LVLMs would require extensive high-quality data and substantial GPU resources, while GPT-based agents would rely on proprietary models (e.g., GPT-4o).
In this paper, we propose \textbf{\underline{Video}} \textbf{\underline{R}}etrieval-\textbf{\underline{A}}ugmented \textbf{\underline{G}}eneration (\textbf{{Video-RAG}}), a training-free and cost-effective pipeline that
employs visually-aligned auxiliary texts to help facilitate cross-modality alignment while providing additional information beyond the visual content.
Specifically, we leverage open-source external tools to extract visually-aligned information from pure video data (e.g., audio, optical character, and object detection), and incorporate the extracted information into an existing LVLM as auxiliary texts, alongside video frames and queries, in a plug-and-play manner. 
Our \textbf{{Video-RAG}} offers several key advantages: (\romannumeral 1) lightweight with low computing overhead due to single-turn retrieval; (\romannumeral 2) easy implementation and compatibility with any LVLM; and (\romannumeral 3) significant, consistent performance gains across long video understanding benchmarks, including Video-MME, MLVU, and LongVideoBench.
Notably, our model demonstrates superior performance over proprietary models like Gemini-1.5-Pro and GPT-4o when utilized with a 72B model. Codes are available at \url{https://github.com/Leon1207/Video-RAG-master}.
  
\end{abstract}

\section{Introduction}
\label{sec:intro}

\begin{figure}[t!]
  \centering 
  \includegraphics[width=1.0\linewidth]{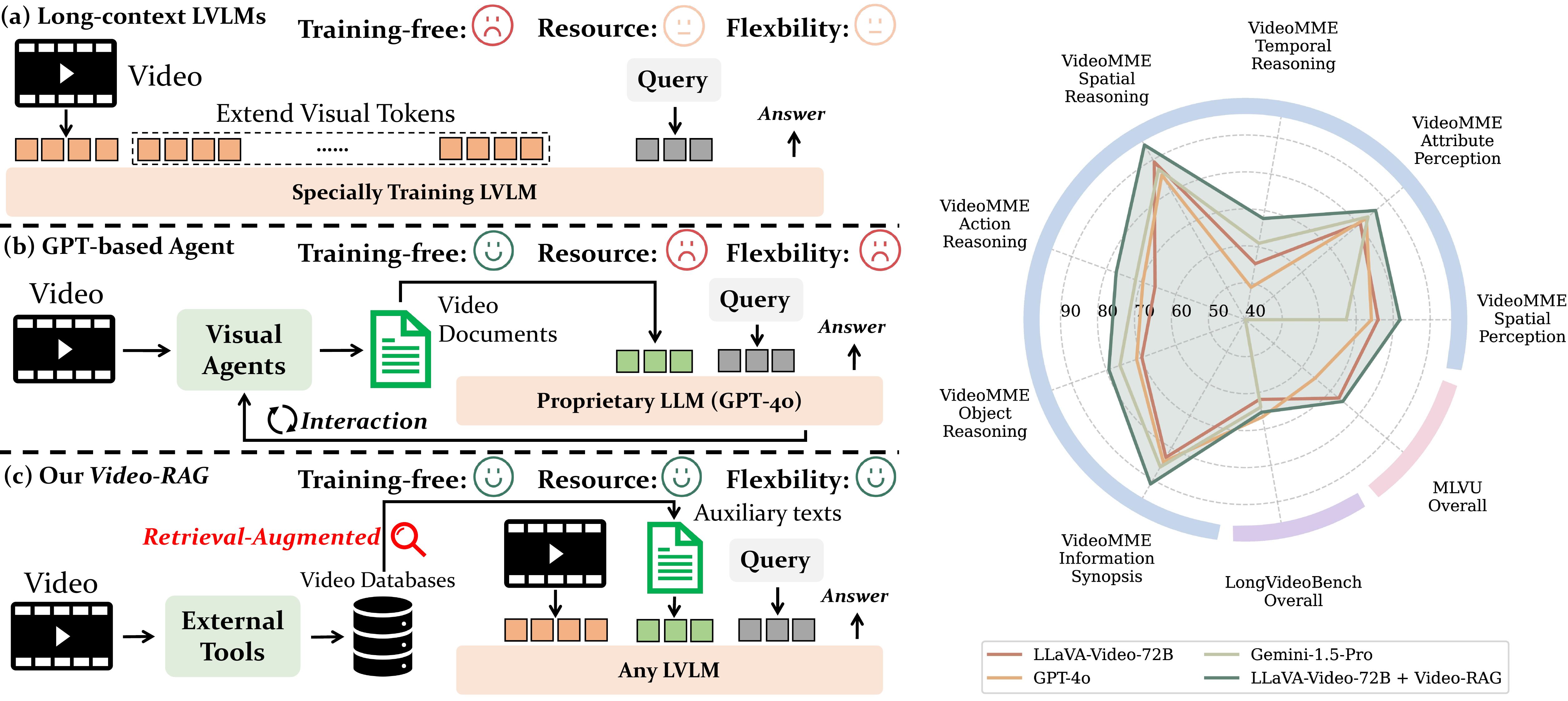}
  \caption{(Left) Two common approaches for understanding long videos, alongside our Video-RAG. Video-RAG provides a resource-efficient, training-free pipeline compatible with any LVLM. By leveraging RAG, it retrieves auxiliary texts for input, leading to notable performance enhancement. (Right) Comparison of the performance of Video-RAG with LLaVA-Video-72B \cite{llavavideo}, Gemini-1.5-Pro \cite{gemini}, and GPT-4o \cite{gpt4o} across various benchmarks, including the sub-tasks from Video-MME \cite{videomme} (we focus only on those that outperform Gemini-1.5-Pro), LongVideoBench \cite{lvb}, and MLVU \cite{mlvu}.}
  \label{fig_intro} 
  \vspace{-15pt}
\end{figure}

With the advancements in Large Language Models (LLMs), numerous studies have been conducted to enhance their ability to comprehend and process videos \cite{chatunivi, videochat, videollava, llavanextvideo, stllm, videollama, vila, internvl, chen2024sharegpt4video, zhang2024beyond, kangaroogroup, liu2024oryx, li2025llama}, collectively termed Large Video-Language Models (LVLMs). Although current LVLMs have demonstrated promising performance in understanding short videos, effective comprehension of extremely long videos continues to be a major challenge.

To address this challenge, recent studies \cite{longva, longllava, intp, wang2024longllava, zong2024long} have sought to extend the reasoning context length of LVLMs, essentially finetuning long-context LVLMs for long video understanding. LongVA \cite{longva} first introduces increasing the token capacity of an LLM and transferring its long-context comprehension capabilities to video data. 
However, training such a model requires pre-training on an extended corpus, and often there are distribution shifts between deployment videos and finetuning videos.
As demonstrated in Video-MME \cite{videomme}, LongVA declines when increasing the video frame sampling rate from 128 to 384 (52.6\% → 51.8\%). This outcome suggests that simply increasing the number of sampled frames not only leads to information redundancy but also imposes additional challenges for the model to handle complex reasoning.
Retrieval-Augmented Generation \cite{rag} (RAG) is a technique that enhances generative tasks by retrieving relevant documents from an external corpus, thus improving response quality in LLMs. Recent studies have begun exploring the integration of RAG with video-based tasks \cite{irag, drvideo, omagent, videorag}, employing tools to process videos in long contexts and sending them to a proprietary model for generation, which is known as the GPT-based Agent method.
However, they come with serval limitations.
First, most of them process long video content as plain text, subsequently utilizing the RAG mechanisms to retrieve relevant documents for LLMs. Therefore, they lack alignment with the visual context of the video, resulting in a loss of critical visual information.
Second, they are often resource-intensive in multi-turn interactions and typically require powerful LLMs to function as the driving force, thus limiting their flexibility and generative capabilities. Executing the whole Video-MME \cite{videomme} using VideoAgent \cite{videoagent} requires approximately 20 days and incurs a substantial consumption of GPT-4o API tokens.

In this study, we propose Video-RAG, an effective RAG pipeline that can be seamlessly integrated with any LVLM.
Specifically, instead of simply increasing the number of sampled video frames, we propose to replace the corresponding extended visual tokens with auxiliary texts extracted from pure video data by invoking open-source foundation models, such as optical character recognition (OCR), automatic speech recognition (ASR), and object detection. These auxiliary texts are more aligned with the visual context while providing additional information beyond the visual data, as demonstrated in \cite{mmvid, videoagent}.
Besides dealing with the context windows limit of LVLMs, we employ RAG in Video-RAG to filter auxiliary texts, ensuring their relevance to the user's query in the text embedding space.
As sampled visual context often lacks explicit alignment with the instructions, the auxiliary texts can facilitate cross-modality alignment while reducing the modality divide. As illustrated in Figure \ref{fig_tsne}, with Video-RAG, the retrieved auxiliary texts help guide the LVLM to pay more attention to the query-relevant keyframes, while simultaneously facilitating cross-modality alignment between query and keyframes.
In this framework, an LVLM serves as the central component of Video-RAG, processing visual tokens to preserve detailed visual context and minimize potential information loss. Moreover, the retrieval process is parallelly executed in a single operation, ensuring efficiency.

We evaluate Video-RAG across several long video benchmarks, including Video-MME \cite{videomme}, MLVU \cite{mlvu}, and LongVideoBench \cite{lvb}. By applying the Video-RAG to seven distinctive open-source LVLMs, we achieve an average performance improvement of 2.8\% on Video-MME with only 2.0K text tokens addition (equal to 14 frames in most configuration) per case, while beating the proprietary LVLM when integrated with the 72B model, as shown in the right part of Figure \ref{fig_intro}.
Applying Video-RAG to a 7B LVLM only requires an additional 8GB of inference GPU memory and approximately 5 seconds of inference time per case (details in the Supplemental Material).

In summary, our contributions are as follows:

\begin{itemize}
    \item \textbf{We integrate RAG into open-source LVLMs:} Video-RAG incorporates three types of visually-aligned auxiliary texts (OCR, ASR, and object detection) processed by external tools and retrieved via RAG, enhancing the LVLM. It's implemented using completely open-source tools, without the need for any commercial APIs.
    \item \textbf{We design a versatile plug-and-play RAG-based pipeline for any LVLM:} Video-RAG offers a training-free solution for a wide range of LVLMs in a plug-and-play manner, delivering performance improvements with minimal additional resource requirements.
    \item \textbf{We achieve proprietary-level performance with open-source models:} Applying Video-RAG to a 72B open-source model yields proprietary-level performance, surpassing models such as GPT-4o and Gemini-1.5-Pro.
\end{itemize}

\section{Related Work}
\label{sec:related_work}

\subsection{Large Video-Language Models}

With the rapid advancement of large language models (LLMs), there has been increasing interest in developing generalist video models capable of handling video-related tasks. Video-ChatGPT \cite{videochatgpt} extracts features from individual frames and aggregates them through both spatial and temporal pooling operations. 
VideoChat \cite{videochat} encodes videos by generating both textual descriptions and video appearance embeddings.
Video-LLaVA \cite{videollava} aligning image and video encoders during a pre-processing phase, using a shared projector to map the encoded representations into a common language space.
LLaVA-NeXT-Video \cite{llavanextvideo} extends LLaVA-NeXT \cite{llavanext} by fine-tuning the model specifically on video data.
Despite their contributions, these approaches face challenges when processing long videos, primarily due to the limited number of frames sampled for analysis.


\subsection{Long-context Large Video-Language Models}

Recent approaches have sought to expand the context window size to enhance long video understanding.
LongVA \cite{longva} and Long-LLaVA \cite{longllava} address this by continuously training LLMs on extended textual data, to transfer their long-text comprehension capabilities to video processing.
INTP \cite{intp} introduces a video token rearrangement technique while proposing a training-free method for extending the LLM context window, allowing LVLMs to process increased visual tokens.
However, these methods face challenges in striking a balance between the high computational costs associated with sampling video frames and the limited performance improvements achieved. Due to the inherent redundancy in video content and constraints on model capacity, performance degradation may occur when the number of sampled frames surpasses a certain threshold.


\subsection{GPT-based Agent Video Understanding}

Initial efforts \cite{zhang2023simple, wang2024videoagent, gupta2023visual, suris2023vipergpt, videorag} have employed LLMs to interact with tools to process visual information as structured long context for question-answering.
MM-VID \cite{mmvid} enhances long video understanding by aligning video frames with corresponding text descriptions.
VLog \cite{vlog} leverages multimodel pre-trained models to capture and interpret visual and audio information, summarizing it into documents for video comprehension.
VideoAgent \cite{videoagent}, DrVideo \cite{drvideo}, and OmAgent \cite{omagent} integrate multimodal inputs and enable dynamic querying of video segments to support long video reasoning tasks.
VideoRAG \cite{videorag} and VideoRAG \cite{jeong2025videorag} achieve a tighter integration between the RAG framework and proprietary models.
However, these methods take a long time to process videos while relying on proprietary models (e.g., GPT-4o), thus limiting their efficiency and adaptability to other open-source frameworks.


\begin{figure*}[t!]
  \centering 
  \includegraphics[width=1.0\linewidth]{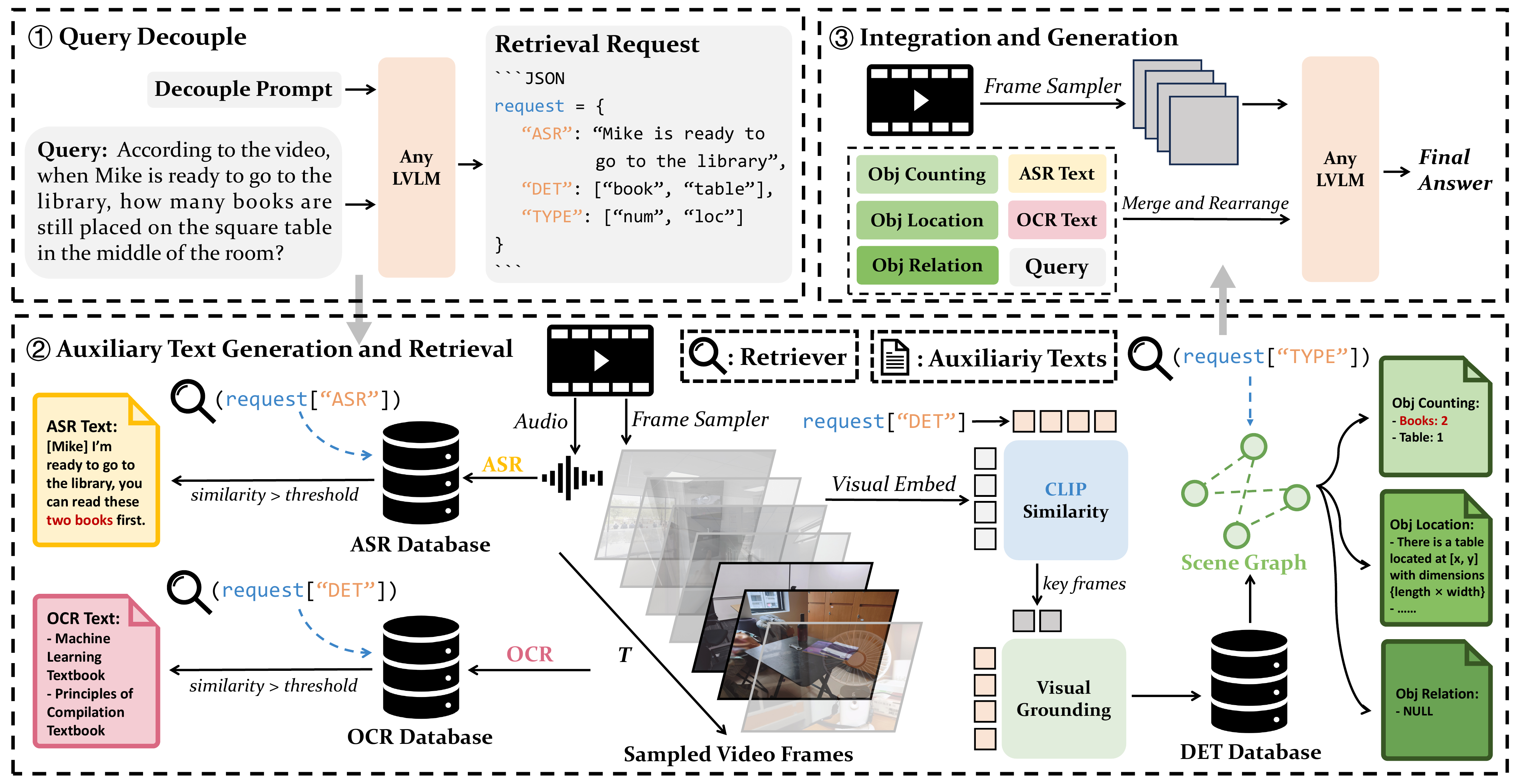}
  \caption{The framework of our Video-RAG. In the query decouple phase, the LVLM is prompted to generate a retrieval request for auxiliary texts. Next, in the auxiliary text generation and retrieval phase, the video is processed \textbf{in parallel} to extract three types of textual information (OCR, ASR, and object detection), and the relevant text is retrieved as the auxiliary text. Finally, in the integration and generation phase, auxiliary texts are combined with the query and the video to generate the response.}
  \label{fig_framework} 
  \vspace{-10pt}
\end{figure*}

\section{Method}
\label{sec:formatting}

We propose a novel, training-free pipeline for large video-language models (LVLMs), named Video-RAG, which can be integrated into any LVLM. As illustrated in Figure \ref{fig_framework}, our pipeline comprises three key phases: \textbf{(\romannumeral 1) Query Decouple: } In this phase, the user’s query is decomposed into a retrieval request aimed at extracting auxiliary texts from the target video. \textbf{(\romannumeral 2) Auxiliary Text Generation \& Retrieval: } Multiple auxiliary texts are generated from the queried video in parallel. Then, the retrieval request is used to obtain relevant external information. \textbf{(\romannumeral 3) Integration and Generation: } This phase integrates the retrieved auxiliary texts with the user’s query, feeding this combined input into the LVLMs to generate the final response.

\subsection{Large Video-Language Model}

Given a video $\bm{\mathrm{V}}$, a frame sampler first sample $N$ frames $\bm{\mathrm{F}}$. Most existing methods uniformly sample frames from a video for both effectiveness and simplicity. Then, video features are extracted as $\bm{\mathrm{F}_v} = \texttt{VisualEnc}(\bm{\mathrm{F}})$, where $\texttt{VisualEnc}$ is an image-based visual encoder, such as CLIP-L \cite{clip}. Finally, the video features $\bm{\mathrm{F}_v}$ and the user's query $\bm{\mathrm{Q}}$ are fed into the LVLM to generate an output $\bm{\mathrm{O}}$:

\begin{equation}
\bm{\mathrm{O}} = \texttt{LVLM}(\bm{\mathrm{F}_v}, \bm{\mathrm{Q}})
\end{equation}

\subsection{Query Decouple}

In this phase, upon receiving a user's query about the video, the LVLM begins by decoupling the query and generating retrieval requests, denoted as $\bm{\mathrm{R}}$, for auxiliary texts. During this phase, the LVLM processes only textual information, without access to video frames, and the output requests are formatted in JSON. We prompt the LVLM using a decoupling prompt $\bm{\mathrm{P}}$ to generate the following retrieval requests as necessary: (\romannumeral 1) $\bm{\mathrm{R}}_{asr}$: Requests about automatic speech recognition, to extract audio information from the video that may pertain to the query. (\romannumeral 2) $\bm{\mathrm{R}}_{det}$: Requests for identifying physical entities within the video that may assist in answering the query. (\romannumeral 3) $\bm{\mathrm{R}}_{type}$: Requests for details about the location, quantity, and relationships of the identified physical entities. These requests, which may be NULL (the corresponding information is not required), are then parsed and forwarded to the auxiliary text retrieval phase. The entire process can be described as:

\begin{equation}
\bm{\mathrm{R}} = \texttt{LVLM}(\bm{\mathrm{P}}, \bm{\mathrm{Q}}), \ \ \bm{\mathrm{R}} = \{\bm{\mathrm{R}}_{asr}, \bm{\mathrm{R}}_{det}, \bm{\mathrm{R}}_{type}\}
\end{equation}

\subsection{Auxiliary Text Generation}

In this phase, we first generate the auxiliary texts from the video and then retrieve them to assist the LVLMs according to the retrieval requests $\bm{\mathrm{R}}$.
As the length of the video increases, the number of tokens generated from the processed data also grows, leading to an increase in redundant information. Additionally, current open-source models are constrained by the limited length of their context windows, which may prevent them from fully processing all auxiliary texts. To address this issue, we draw inspiration from Retrieval-Augmented Generation (RAG) \cite{rag}, retrieving only the auxiliary texts relevant to the user's query.
Before retrieval, we construct the necessary databases from the given video in parallel. Specifically, we implement three distinct databases: the Optical Character Recognition (OCR) database, denoted as $DB_{ocr}$; the Automatic Speech Recognition (ASR) database, denoted as $DB_{asr}$; and the Object Detection (DET) database, denoted as $DB_{det}$.

\noindent \textbf{OCR database.} Current LVLM are still illusory in their ability to accurately recognize characters, and their performance often falls short compared to proprietary models. To better leverage the information contained in video frames and reduce hallucinations, we employ a proprietary OCR model to extract text from each video frame with the same frame-sampled strategy as LVLMs. Specifically, we use EasyOCR \cite{easyocr} as our text recognition model and segmented the recognized texts on a per-frame basis, denoted as $\bm{\mathrm{T}}_{ocr}$. Subsequently, we implemented RAG by utilizing the advanced text encoding model Contriever \cite{contriever} to encode the fetched OCR texts into text embeddings $\bm{\mathrm{E}}_{ocr}$. These embeddings are then stored in a database with the FAISS index \cite{faiss}, a library designed for efficient similarity search and clustering of dense vectors. The entire building process can be formally described as:
\begin{equation}
\bm{\mathrm{T}}_{ocr} = \texttt{EasyOCR}(\bm{\mathrm{F}})
\end{equation}
\begin{equation}
DB_{ocr} \xleftarrow{\ \texttt{FAISS}\ } \bm{\mathrm{E}}_{ocr} = \texttt{Contriever}(\bm{\mathrm{T}}_{ocr}) 
\end{equation}

\noindent \textbf{ASR database.} Audio information (e.g., subtitles) plays a crucial role in video comprehension, often providing additional context that may not be available through visual cues alone. To incorporate them, we first extract the raw audio $\bm{\mathrm{U}}$ from the video and then transcribe them into texts $\bm{\mathrm{T}}_{asr}$. Specifically, we use Whisper \cite{whisper} as our audio transcription model. Since the recognized texts can be quite extensive, we chunk and encode them into a vector database, following the same procedure used to construct the OCR database. The building process can be formally described as:

\begin{equation}
\bm{\mathrm{T}}_{asr} = \texttt{Whisper}(\bm{\mathrm{U}})
\end{equation}
\begin{equation}
DB_{asr} \xleftarrow{\ \texttt{FAISS}\ } \bm{\mathrm{E}}_{asr} = \texttt{Contriever}(\bm{\mathrm{T}}_{asr})
\end{equation}

\noindent \textbf{DET database.} While LVLMs demonstrate strong performance in object recognition, they continue to face challenges such as object counting, precise object localization, and understanding relative relationships between objects. To mitigate the issue of hallucination, which can stem from these challenges, we incorporate object detection information as auxiliary texts.
We leverage a visual grounding model to extract both the object categories and their corresponding positions from sampled video frames. This approach helps provide more accurate and context-aware object detection. 
To enhance processing efficiency, we limit object detection to keyframes only. Specifically, we compute the CLIP similarity \cite{clip} between the object retrieval request $\bm{\mathrm{R}}_{det}$ and the sampled video frames $\bm{\mathrm{F}}$ and select relevant keyframes $\bm{\mathrm{F}}_{key}$ based on a threshold $t$:

\begin{equation}
\bm{\mathrm{F}}_{key} = \texttt{CLIP\_similarity}(\bm{\mathrm{R}}_{det}, \bm{\mathrm{F}}) > t
\end{equation}

Once the keyframes are identified, we utilize APE \cite{ape}, an efficient open-vocabulary object detection model that accepts object descriptions as prompts to detect relevant objects within frames based on retrieval queries. The capability of APE makes it particularly well-suited to our requirements for on-demand object retrieval. Finally, the detected objects' categories and their corresponding positional information are stored in the DET database using natural language representations:

\begin{equation}
DB_{det} \xleftarrow{} \bm{\mathrm{T}}_{det} = \texttt{APE}(\bm{\mathrm{F}}_{key}, \bm{\mathrm{R}}_{det})
\end{equation}

\subsection{Auxiliary Text Retrieval}

During the retrieve phase, we employ Contriever \cite{contriever} to encode the user's query and the parsed requests for OCR and ASR into text embeddings, then concatenating to form the final query request $\bm{\mathrm{E}}_{req} = \texttt{Contriever}(\texttt{Concat}(\bm{\mathrm{R}}, \bm{\mathrm{Q}})), \ \bm{\mathrm{R}} \in \{\bm{\mathrm{R}}_{ocr}, \bm{\mathrm{R}}_{asr}\}$. Then we retrieve the auxiliary texts from $DB \in \{DB_{ocr}, DB_{asr}\}$ by the FAISS tool, which computes the vector similarity between the query and text chunks stored in the database. Text chunks with a FAISS similarity score greater than threshold $t$ are indexed as the retrieval results $\bm{\mathrm{A}} \in \{\bm{\mathrm{A}}_{ocr}, \bm{\mathrm{A}}_{asr}\}$. The process can be formulated as:

\begin{equation}
\bm{\mathrm{A}} \xleftarrow{\texttt{Index}} \texttt{FAISS\_similarity}(DB, \bm{\mathrm{E}}_{req}) > t
\end{equation}

Since the text generated by the detection model is in a raw format (``category: [x\_min, y\_min, length, width]"), it challenges LVLMs to understand the relative relationships between objects. We preprocess the object information using a scene graph, which helps to represent spatial and relational information more explicitly. This preprocessing allows us to construct more coherent and semantically meaningful texts, denoted as $\bm{\mathrm{A}}_{det}^{p}$, which are more readily interpretable by LVLMs.
We incorporate three types of object information for each video keyframe:
\textbf{(\romannumeral 1) Object Location $\bm{\mathrm{A}}_{loc}$:} This refines the positional information of the object, formatted as: ``Object \{node ID\} is a \{object category\} located at coordinates [x, y] with dimensions \{length $\times$ width\}''
\textbf{(\romannumeral 2) Object Counting $\bm{\mathrm{A}}_{cnt}$:} This counts the number of objects and generates text in the following format: ``Object counting: - \{object category\}: \{number\}'' 
\textbf{(\romannumeral 3) Relative Positional Relationships $\bm{\mathrm{A}}_{rel}$:} This captures the relative spatial relationships between objects using the format: ``Object \{node ID\} (\{object category\}) is \textless{}positional description\textgreater{} Object \{node ID\} (\{object category\})''. 
By combining this information, we construct a detailed representation of the objects in the frame, denoted as $\bm{\mathrm{A}}_{det}^{p} = 
\{\bm{\mathrm{A}}_{loc}, \bm{\mathrm{A}}_{cnt}, \bm{\mathrm{A}}_{rel}\}$:

\begin{equation}
\bm{\mathrm{A}}_{det}^{p} = \texttt{SceneGraph}(DB_{det})
\end{equation}

Finally, we acquire the object auxiliary texts based on the object information type retrieval requests $\bm{\mathrm{R}}_{type}$, which selects and finalizes the object auxiliary information $\bm{\mathrm{A}}_{det}$. $\bm{\mathrm{A}}_{det}$ is one of the elements of the power set $\mathcal{P}$ of $\bm{\mathrm{A}}_{det}^{p}$ selected by $\bm{\mathrm{R}}_{type}$, and the retrieve process can be formulated as:

\begin{equation}
\bm{\mathrm{A}}_{det} = \bm{\mathrm{R}}_{type}(\mathcal{P}(\bm{\mathrm{A}}_{det}^{p})) \in \mathcal{P}(\bm{\mathrm{A}}_{det}^{p})
\end{equation}

\subsection{Integration and Generation}

After obtaining different types of auxiliary texts, we organize them chronologically using natural language to create a unified auxiliary input, denoted as $\bm{\mathrm{A}}_{m} = \texttt{Concat}(\bm{\mathrm{A}}_{ocr}, \bm{\mathrm{A}}_{asr}, \bm{\mathrm{A}}_{det})$. These merged auxiliary inputs, along with the user's query and the sampled video frames, are then fed into the LVLM to produce the final result. The overall process can be formulated as:

\begin{equation}
\bm{\mathrm{O}} = \texttt{LVLM}(\bm{\mathrm{F}_v}, \texttt{Concat}(\bm{\mathrm{A}}_{m}, \bm{\mathrm{Q}}))
\end{equation}

\section{Experiments}
\label{sec:exp}

\subsection{Datasets}

\noindent \textbf{Video-MME} \cite{videomme} is a widely used benchmark for assessing the ability of LVLMs to handle detailed videos in real-world scenarios. It is divided into three subsets based on video length, with durations ranging from 11 seconds to 1 hour. 
\textbf{MLVU} \cite{mlvu} is a long video understanding benchmark with a large wide of 9 distinct tasks. It is created based on long videos of diversified lengths, ranging from 3 minutes to 2 hours with about 12 minutes average video length.
\textbf{LongVideoBench} \cite{lvb} is a benchmark designed to accurately retrieve and reason over detailed multimodal information from long videos, with 6,678 human-annotated multiple-choice questions in 17 fine-grained categories. 

\begin{table*}[t!]
\centering
\caption{Performance on the Video-MME \cite{videomme} benchmark in without subtitles (w/o S), with subtitles (w/ S) and equipped with our Video-RAG (Ours), \textbf{Frames} and \textbf{Gain} means the input frame number and performance gain by applying Video-RAG compared to the baseline with subtitles. By applying Video-RAG to seven LVLMs, we observed an average performance improvement of 2.8\% by adding only an average of $\sim$2.0K auxiliary texts compared to $\sim$3.0K full-subtitled tokens per sample. In particular, we perform better when applying Video-RAG with 72B LLaVA-Video \cite{llavavideo} than the proprietary method GPT-4o \cite{gpt4o} (77.4\% vs. 77.2\%). All results are our republication.}
\vspace{3pt}
\label{tab_videomme}
\setlength{\tabcolsep}{0.4mm}
\renewcommand\arraystretch{1.0} 
\scalebox{0.84}[0.84]{
\begin{tabular}{l|cc|cc>{\columncolor{cyan!10}}c|cc>{\columncolor{cyan!10}}c|cc>{\columncolor{cyan!10}}c|cc>{\columncolor{cyan!10}}c|c}
\toprule
\multirow{2}{*}{\textbf{Model}}   & \multirow{2}{*}{\textbf{Params}} & \multirow{2}{*}{\textbf{Frames}} & \multicolumn{3}{c|}{\textbf{Short}} & \multicolumn{3}{c|}{\textbf{Medium}} & \multicolumn{3}{c|}{\textbf{Long}} & \multicolumn{3}{c|}{\textbf{Overall}} & \multirow{2}{*}{\textbf{Gain}}  \\
 &  &  & w/o S & w/ S & Ours & w/o S & w/ S & Ours & w/o S & w/ S & Ours & w/o S & w/ S & Ours & \\
  \midrule
    \multicolumn{16}{c}{\textbf{\textit{Proprietary LVLMs}}} \\
\midrule
\rowcolor{gray!20} GPT-4o \cite{gpt4o} & - & 384 & 80.0 & 82.8 & - & 70.3 & 76.6 & - & 65.3 & 72.1 & - & 71.9 & 77.2 & - & - \\
\rowcolor{gray!20} Gemini-1.5-Pro \cite{gemini} & - & 0.5 fps & 81.7 & 84.5 & - & 74.3 & 81.0 & - & 67.4 & 77.4 & - & 75.0 & 81.3 & - & - \\
  \midrule
      \multicolumn{16}{c}{\textbf{\textit{Open-Source LVLMs}}} \\
      \midrule
Video-LLaVA \cite{videollava} & 7B & 8 & 45.3 & 46.1 & 49.5 & 38.0 & 40.7 & 43.0 & 36.2 & 38.1 & 42.5 & 39.9 & 41.6 & 45.0 & \textcolor{red}{+3.4} \\ 
LLaVA-NeXT-Video \cite{llavanextvideo} & 7B & 16 & 49.4 & 51.8 & 56.6 & 43.0 & 46.4 & 47.4 & 36.7 & 44.9 & 46.0 & 43.0 & 47.7 & 50.0 & \textcolor{red}{+2.3} \\ 
VITA-1.5 \cite{vita15} & 7B & 16 & 67.0 & 69.9 & 71.0 & 54.2 & 55.7 & 55.4 & 47.1 & 50.4 & 52.4 & 56.1 & 58.7 & 59.6 & \textcolor{red}{+0.9} \\ 
LongVA \cite{longva} & 7B & 128 & 61.1 & 61.2 & 66.1 & 50.4 & 53.8 & 60.4 & 46.2 & 52.9 & 59.4 & 52.6 & 56.0 & 62.0 & \textcolor{red}{+6.0}\\ 
Long-LLaVA \cite{longllava} & 7B & 64 & 61.9 & 62.4 & 67.1 & 51.4 & 56.2 & 60.4 & 45.4 & 54.7 & 60.1 & 52.9 & 57.8 & 62.6 & \textcolor{red}{+4.8} \\
Qwen2-VL \cite{qwen2vl} & 72B & 32 & 75.0 & 76.7 & 77.4 & 63.3 & 69.9 & 70.2 & 56.3 & 69.2 & 71.0 & 64.9 & 71.9 & 72.9 & \textcolor{red}{+1.0} \\ 
LLaVA-Video \cite{llavavideo}  & 72B & 64 & 80.7 & 81.8 & 82.8 & 68.7 & 73.8 & 76.3 & 62.1 & 72.2 & 73.1 & 70.3 & 75.9 & 77.4 & \textcolor{red}{+1.5} \\  

  \bottomrule
\end{tabular}}
\label{tab_compare}
\vspace{-10pt}
\end{table*}

\subsection{Implementation Details}

We performed all experiments on NVIDIA A100 80G GPUs. 
During the auxiliary text generation phase, we first restrict the detection requests $\bm{\mathrm{R}}_{det}$ generated by LVLMs in decouple prompt then further filter them using spaCy, ensuring they correspond to CLIP-sensitive physical entities, avoiding the inclusion of abstract concepts. 
In the auxiliary text retrieval phase, we set both the CLIP similarity threshold and the FAISS similarity threshold $t$ to 0.3.
We employ the IndexFlatIP as the similarity calculating method of FAISS \cite{faiss}. 
Note that we don't include the GPT-based Agent methods for comparison due to their resource-intensive nature (complete execution of Video-MME \cite{videomme} costs around \$2000 for API purchasing when using VideoAgent \cite{videoagent}). Still, we include a mini-experiment of VideoAgent in the Supplemental Material that compares the overall performance, inference time, and GPU requirements with two common long-context LVLMs and our Video-RAG.

\begin{table}[ht]
\centering
\resizebox{1.0\textwidth}{!}{ 
    \begin{minipage}[c]{0.5\linewidth}
        \centering
        \setlength{\tabcolsep}{0.6mm} 
        \renewcommand{\arraystretch}{1.0}
        \caption{The overall performance in the multiple-choice task of the MLVU \cite{mlvu} benchmark. * donates the results of our replication.}
        \scalebox{0.84}[0.84]{
        \begin{tabular}{lcc|c}
\toprule

\textbf{Model} & \textbf{\#Params}  & \textbf{Frames} & \textbf{Overall}   \\ \midrule
\multicolumn{4}{c}{\textbf{\textit{Proprietary LVLMs}}} \\ \midrule
\rowcolor{gray!20} GPT-4o \cite{gpt4o} & - & 0.5 fps & 64.6   \\   \midrule
\multicolumn{4}{c}{\textbf{\textit{Open-Source LVLMs}}} \\   \midrule
VITA-1.5 \cite{vita15} & 7B & 16 & 60.4  \\
Video-CCAM \cite{videoccam} & 14B & 96 & 63.1  \\
Video-XL \cite{videoxl} & 7B & 256 & 64.9  \\
Aria \cite{aria} & 25.3B & 256 & 70.6   \\
LLaVA-Video* \cite{llavavideo} & 7B & 64 & 70.8 \\
Oryx-1.5 \cite{oryx} & 32B & 128 & 72.3 \\
LLaVA-Video* \cite{llavavideo} & 72B & 64 & \underline{73.1} \\ \midrule
\rowcolor{cyan!10} LLaVA-Video + Video-RAG & 7B & 64 & 72.4 \\
\rowcolor{cyan!10} LLaVA-Video + Video-RAG & 72B & 64 & \textbf{73.8} \\
\bottomrule
\end{tabular}}
\label{tab_mlvu}
    \end{minipage}
    \hspace{0.2cm}
    \begin{minipage}[c]{0.51\linewidth}
        \centering
        \setlength{\tabcolsep}{0.6mm} 
        \renewcommand{\arraystretch}{1.0}
        \caption{The overall performance on the validation set of LongVideoBench \cite{lvb}. * donates the results of our replication.}
        \scalebox{0.84}[0.84]{
        \begin{tabular}{lcc|c}
\toprule

\textbf{Model} & \textbf{\#Params}  & \textbf{Frames} & \textbf{Overall}   \\ \midrule
\multicolumn{4}{c}{\textbf{\textit{Proprietary LVLMs}}} \\ \midrule
\rowcolor{gray!20} Gemini-1.5-Pro \cite{gemini} & - & 256 & 64.0   \\ 
\rowcolor{gray!20} GPT-4o \cite{gpt4o} & - & 256 & \textbf{66.7}   \\ 
 \midrule
\multicolumn{4}{c}{\textbf{\textit{Open-Source LVLMs}}} \\   \midrule
VideoChat2-Mistral \cite{videochat} & 7B & 8 & 39.3  \\
ShareGPT4Video \cite{chen2024sharegpt4video} & 7B & 8  & 39.7   \\
LLaVA-Next-Mistral \cite{llavanext} & 7B & 8 & 49.1   \\ 
PLLaVA \cite{pllava} & 34B & 16 & 53.2   \\ 
VITA-1.5 \cite{vita15} & 7B & 16 & 53.6  \\
LLaVA-Video* \cite{llavavideo} & 7B & 64 & 56.6   \\ 
LLaVA-Video* \cite{llavavideo} & 72B & 64 & 61.9  \\ \midrule
\rowcolor{cyan!10} LLaVA-Video + Video-RAG & 7B & 64 & 58.7 \\
\rowcolor{cyan!10} LLaVA-Video + Video-RAG & 72B & 64 & \underline{65.4} \\

\bottomrule
\end{tabular}}
\label{tab_lvb}
    \end{minipage}
    }
    \vspace{-10pt}
\end{table}

\subsection{Main Results}

\textbf{Video-MME}. We evaluate our Video-RAG in five 7B open-source LVLMs, including Video-LLaVA \cite{videollava}, LLaVA-NeXT-Video \cite{llavanextvideo}, LongVA \cite{longva}, Long-LLaVA \cite{longllava}, and two 72B LVLM Qwen2-VL \cite{qwen2vl} and LLaVA-Video \cite{llavavideo}. Constraining by computational resources, we evaluate the LVLMs with their official frame rate setting in Video-MME except for 72B Qwen2-VL, which requires about 3K GPU memory with 768 video frame input ($\sim$38 A100 GPUs).
Results are shown in Table \ref{tab_compare}. Specifically, after applying our Video-RAG in 72B LLaVA-Video \cite{llavavideo}, we perform better than the proprietary model GPT-4o \cite{gpt4o} (77.4\% vs. 77.2\%). Across the seven LVLMs used in our experiments, we gained an average performance boost of 2.8\% compared to results with subtitles, especially a significant gain on long videos, demonstrating its effectiveness. This performance improvement is achieved by incorporating token counts from approximately 14 additional video frames (equivalent to 2.0K tokens), each contributing around 144 tokens under most LVLM configurations.
We obtain such a large performance enhancement because most LVLMs are pre-trained primarily within the text space and aligned with visual information, often lacking explicit alignment between embedding spaces. Auxiliary texts can serve as semantic supplements sensitive to LVLMs, facilitating model activation and easing the understanding of complex videos.

\noindent \textbf{MLVU}. We evaluate Video-RAG when integrating into the 7B and 72B LLaVA-Video \cite{llavavideo} of MLVU \cite{mlvu}, a benchmark that is close to performance saturation. As shown in Table \ref{tab_mlvu}, Video-RAG's 1.6\% improvement at 7B-scale is substantial, considering that it outperforms the 32B Qryx-1.5 \cite{oryx} by 0.1\%, while recent 7B-scale models average only a 1.3\% gain (across 15 approaches in MLVU's leaderboard). Additionally, the 72B LLaVA-Video also has a performance gain of 0.7\%, which sets a new state-of-the-art.

\noindent \textbf{LongVideoBench}. We evaluate Video-RAG when applied in the 7B and 72B LLaVA-Video \cite{llavavideo} of LongVideoBench \cite{lvb}. We omit the interleaved input format introduced in LongVideoBench when applying Video-RAG. The evaluation results in Table \ref{tab_lvb} demonstrate that 72B LLaVA-Video with our Video-RAG achieves an overall performance of 65.4\% on the validation set. This result surpasses the proprietary LVLM Gemini-1.5-Pro \cite{gemini} by 1.4\%, securing the second place, just 1.3\% behind GPT-4o \cite{gpt4o}. Meanwhile, the 7B LLaVA-Video also has a performance enhancement of 2.1\% when equipped with our Video-RAG.

\subsection{Ablation Studies}

\begin{wrapfigure}{r}{0.6\textwidth} 
  \centering
  \includegraphics[width=0.95\linewidth]{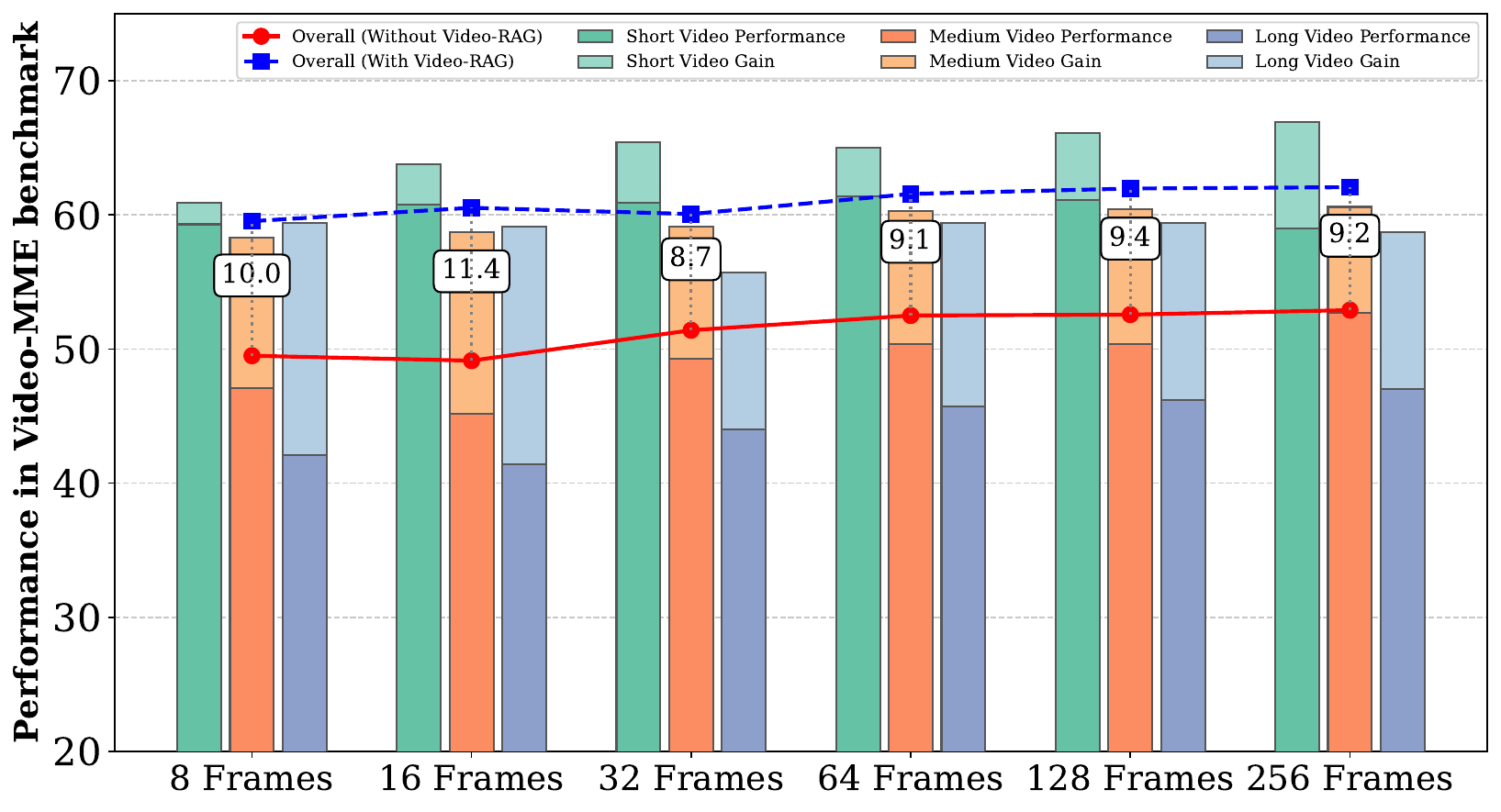} 
  \caption{Performance gain with different sampling frames rate on Video-MME \cite{videomme} when implement LongVA-7B \cite{longllava}.}
  \label{fig_abs_frame}
  \vspace{-10pt} 
\end{wrapfigure}

\noindent \textbf{Effect of different sampling frame number.} To explore the effect of the number of sampling frames on Video-RAG, we experience sampling frames number of 8, 16, 32, 64, 128, and 256 in 7B model LongVA \cite{longva}, results are shown in Figure \ref{fig_abs_frame}. 
As demonstrated, Video-RAG consistently delivers performance improvements across all frame rates, especially in long videos.
The experimental results also indicate that Video-RAG can achieve higher performance gains with fewer frames, demonstrating its potential for applications under resource-constrained conditions.

\noindent \textbf{Effect of different components of Video-RAG.} To explore the effectiveness of auxiliary texts, we add DET, OCR, and ASR as auxiliary texts before and after retrieving by the RAG to evaluate Long-LLaVA-7B \cite{longllava} with 32-frame setting in the Video-MME \cite{videomme} benchmark. 
As shown in Table \ref{abl_comp}, the performance of Long-LLaVA progressively improves as auxiliary texts after retrieving by the RAG system are incrementally added (52.0\% → 52.9\% → 55.7\% → 62.1\%). Among these components, ASR auxiliary texts contribute to a general improvement for different video durations, especially for long videos. When all components are integrated, we obtain an optimal performance, as shown in the last row of Table \ref{abl_comp}.
Meanwhile, the experiment shows a 2.3\% improvement (59.8\% vs 62.1\%) in performance after incorporating RAG for retrieval, demonstrating that auxiliary texts after retrieving by the RAG system are query-aligned, which helps cross-modality alignment.
We also evaluate across sub-tasks within Video-MME \cite{videomme} and other video benchmarks like MLVU \cite{mlvu}, LongVideoBench\cite{lvb}, and VNBench \cite{vnbench}, more details are shown in the Supplemental Material. 

\begin{table}[ht]
\centering
\resizebox{1.0\textwidth}{!}{ 
    \begin{minipage}[c]{0.5\linewidth}
        \centering
        \setlength{\tabcolsep}{0.6mm} 
        \renewcommand{\arraystretch}{1.0}
        \caption{Results on combinations of different auxiliary texts in Video-MME \cite{videomme} when using Long-LLaVA-7B \cite{longllava} as the LVLM.}
        \scalebox{0.84}[0.84]{
        \begin{tabular}{c|ccc|ccc|c}
\toprule
 \textbf{RAG} & \textbf{DET} & \textbf{OCR} & \textbf{ASR} & \textbf{Short} & \textbf{Medium} & \textbf{Long} & \textbf{Overall}   \\ \midrule
 &  &   &  & 60.3  & 51.4 & 44.1 & 52.0    \\ 
 &   &  & \checkmark  & 62.2 & 55.4 & 54.4 & 57.4     \\ 
 &   & \checkmark & \checkmark  & \underline{64.0} & 56.2 & 55.0 & 58.4    \\ 
 &  \checkmark &  & \checkmark  & 63.0 & \underline{57.3} & \textbf{56.4} & \underline{58.9}    \\ 
\rowcolor{cyan!10} & \checkmark   & \checkmark  & \checkmark  & \textbf{64.3} & \textbf{58.8} & \underline{56.3} & \textbf{59.8} \\ \midrule
 \checkmark & \checkmark  &  & & 61.4 & 51.9 & 45.2 & 52.9    \\ 
 \checkmark &  & \checkmark  &   & 63.2 & 53.2 & 46.3 & 54.3 \\
\checkmark &  & & \checkmark & 65.1 & 59.1 & \textbf{60.7} & 61.6  \\ 
 \checkmark & \checkmark  & \checkmark & & 64.1 & 54.6 & 48.4 & 55.7  \\
 \checkmark & \checkmark  &  & \checkmark & 64.9 & 59.0 & \textbf{60.7} & 61.5    \\
\checkmark &   & \checkmark & \checkmark & \underline{66.3} & \textbf{60.3} & 59.3 & \underline{62.0}    \\ 
\rowcolor{cyan!10} \checkmark & \checkmark   & \checkmark  & \checkmark  & \textbf{66.4} & \underline{60.2} & \underline{59.8} & \textbf{62.1}     \\
  \bottomrule
\end{tabular}
\label{abl_comp}}
\label{tab_mlvu}
    \end{minipage}
    \hspace{0.2cm}
    \begin{minipage}[c]{0.51\linewidth}
        \centering
        \setlength{\tabcolsep}{0.6mm} 
        \renewcommand{\arraystretch}{1.0}
        \caption{Performance with different thresholds of retrieval on Video-MME \cite{videomme} when using Long-LLaVA-7B \cite{longllava} as the LVLM. \textbf{\#Token} and \textbf{Time} denote the total token number of the auxiliary texts and the average inference time per question, respectively.}
        \scalebox{0.86}[0.86]{
        \begin{tabular}{c|cc|ccc|c}
\toprule
\textbf{$t$} & \textbf{\#Token} & \textbf{Time} & \textbf{Short} & \textbf{Medium}  & \textbf{Long} & \textbf{Overall}   \\ \midrule
0.0 & 3.6K & 36s & \textbf{67.6} & 59.4 & 59.1 & \underline{62.0}    \\ 
0.1 & 3.4K & 30s & \underline{67.0} & \underline{59.7} & 59.1 & 61.9    \\ 
0.2 & 2.7K & 18s & 66.0 & \textbf{60.2} & \underline{59.2} & 61.8    \\ 
\rowcolor{cyan!10} 0.3 & 1.9K & 11s & 66.4 & \textbf{60.2} & \textbf{59.8} & \textbf{62.1}   \\ 
0.4 & 0.8K  & 8s &  65.6 & 58.0 & 58.3 & 60.6     \\ 
0.5 & 0.3K  & 7s & 63.1 & 54.9 & 50.2 & 56.1     \\ 
1.0 & 0.0K  & 6s & 60.3 & 51.4 & 44.1 & 52.0     \\ 
\midrule
rnd & 1.9K  & 11s & 65.7 & 55.8 & 56.0 & 59.1     \\ 
\bottomrule
\end{tabular}

\label{abl_thres}}
\label{tab_lvb}
    \end{minipage}
    }
\end{table}

\begin{figure*}[h!]
  \centering 
  \includegraphics[width=0.95\linewidth]{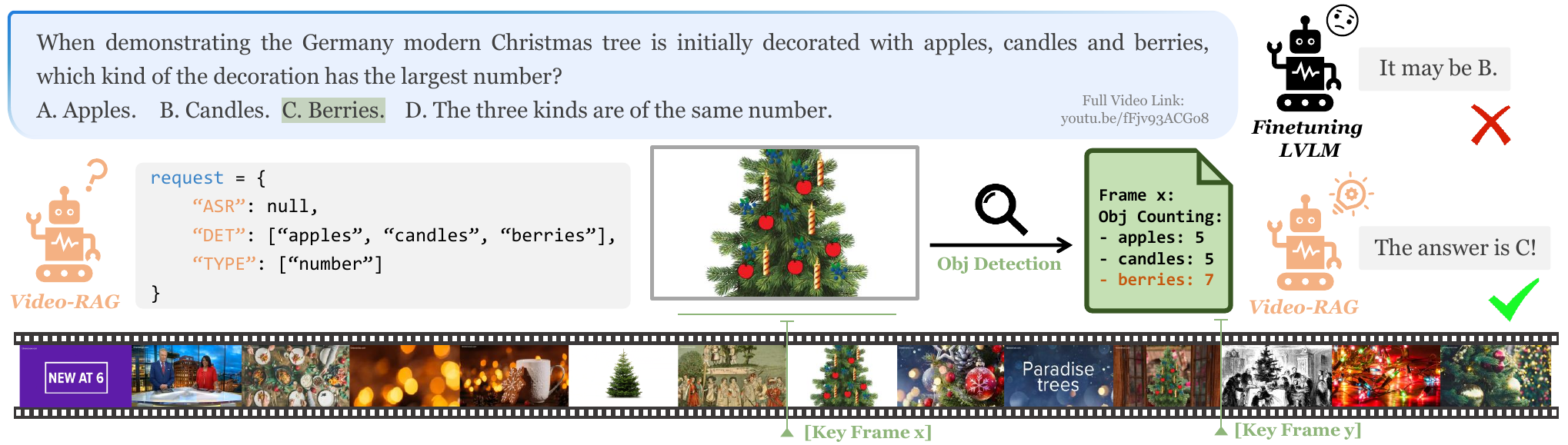}
  \caption{Qualitative result on Video-MME \cite{videomme} when applying Video-RAG with LLaVA-Video \cite{llavavideo}.}
  \label{fig_exam} 
  \vspace{-10pt}
\end{figure*}

\begin{figure*}[h!]
  \centering 
  \includegraphics[width=0.95\linewidth]{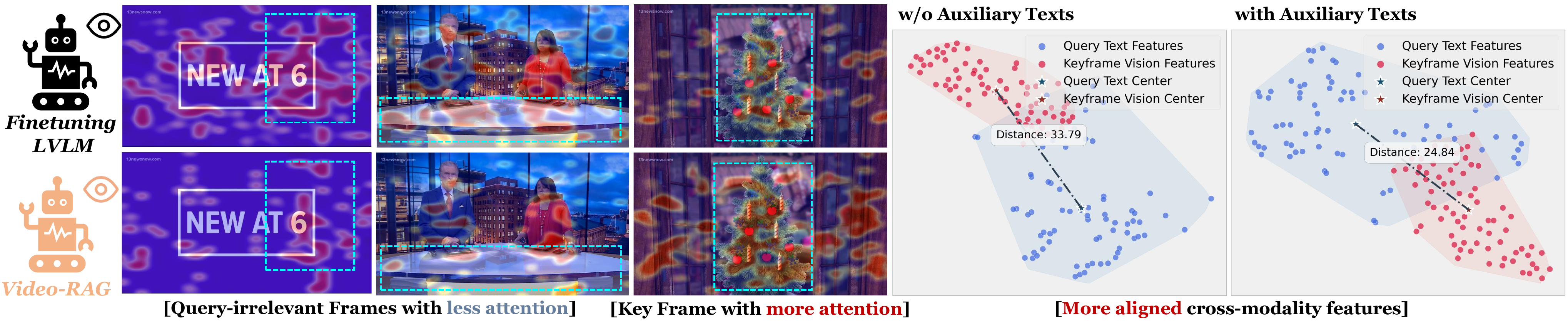}
  \caption{Grad-CAM visualizations of the last hidden state heatmap along with t-SNE visualizations of the user's query and keyframe features of the example shown in Figure \ref{fig_exam}. The retrieved auxiliary texts help \textbf{cross-modality alignment} by assisting the model to \textbf{pay more attention to query-relevant keyframes} and thus generate more robust and accurate answers to the user's query.}
  \label{fig_tsne} 
  \vspace{-5pt}
\end{figure*}

\noindent \textbf{Effect of different thresholds of RAG processing.} When retrieving, we specify a similarity threshold $t$ as a criterion for information selection. 
In the retrieval for OCR and ASR texts, information is selected if its FAISS similarity exceeds $t$. For object detection, frames are selected as keyframes based on their CLIP similarity surpassing $t$, and the relevant information is then extracted. 
Setting $t$ too high may hinder the retrieval of relevant information while setting it too low can result in information redundancy and increased reasoning complexity. To investigate this trade-off, we conduct ablation experiments to evaluate the impact of different threshold values.
The results are shown in Table \ref{abl_thres}. Notably, $t=0$ and $t=1$ correspond to all auxiliary texts input into the model and no auxiliary texts input, respectively. To balance performance with information density and processing time (especially APE \cite{ape} detection in keyframes), we selected a threshold of 0.3 for our implementation. More details about similarity scores are shown in the Supplemental Material. Under this configuration, the additional text length of approximately 1.9K tokens typically remains within the context window limits of open-source LVLMs. For models with more stringent context window limitations, a threshold of 0.4 may also be a viable option. We also randomly sample an equivalent token number of auxiliary texts to serve as inputs for assessing the effectiveness of RAG retrieval, as shown in the last row of Table \ref{abl_thres}.

\subsection{Qualitative Evaluation}

We present qualitative results in the case of Video-MME \cite{videomme} in Figure \ref{fig_exam} and Figure \ref{fig_tsne}. As illustrated, augmenting LLaVA-Video with external tools to process and retrieve auxiliary texts from videos significantly enhances its ability to reduce visual hallucinations, thereby enabling more accurate responses to user queries. Grad-CAM \cite{gradcam} and t-SNE \cite{tsne} visualization results also show that applying Video-RAG helps the LVLM's cross-modality alignment.

\section{Conclusion}
\label{sec:con}

In this paper, we present Video-RAG for effective long video understanding through integrating retrieved auxiliary texts with LVLMs, achieving proprietary-level performance with 72B open-source LVLM. Unlike traditional methods that are resource-intensive with limited gains, Video-RAG offers a resource-efficient, plug-and-play solution leveraging only open-source tools to extract visually-aligned auxiliary texts from video data. However, Video-RAG may be limited by the visual tools we choose and their performance, which lacks adaptation. In the future, we will explore how to more efficiently integrate auxiliary texts and provide an adaptive frame selection strategy for LVLMs.

\section{Acknowledge}

This work was supported by the National Science Fund for Distinguished Young Scholars (No.62025603), the National Natural Science Foundation of China (No. U21B2037, No. U22B2051, No. U23A20383, No. U21A20472, No. 62176222, No. 62176223, No. 62176226, No. 62072386, No. 62072387, No. 62072389, No. 62002305 and No. 62272401), the Natural Science Foundation of Fujian Province of China (No. 2021J06003, No.2022J06001), and the Fundamental Research Funds for the Central Universities.

\bibliographystyle{plainnat} 
\bibliography{neurips_2025}

@misc{vlog,
  author       = {Qinghong Lin},
  title        = {VLog: Transform Video as a Document with ChatGPT, CLIP, BLIP2, GRIT, Whisper, LangChain},
  year         = {2023},
  howpublished = {\url{https://github.com/showlab/VLog}}
}

@misc{easyocr,
  author       = {JaidedAI},
  title        = {EasyOCR},
  year         = {2023},
  howpublished = {\url{https://github.com/JaidedAI/EasyOCR}}
}

@misc{llavanextvideo,
  title={LLaVA-NeXT: A Strong Zero-shot Video Understanding Model},
  url={https://llava-vl.github.io/blog/2024-04-30-llava-next-video/},
  author={Zhang, Yuanhan and Li, Bo and Liu, haotian and Lee, Yong jae and Gui, Liangke and Fu, Di and Feng, Jiashi and Liu, Ziwei and Li, Chunyuan},
  month={April},
  year={2024}
}

@misc{llavanext,
    title={LLaVA-NeXT: Improved reasoning, OCR, and world knowledge},
    url={https://llava-vl.github.io/blog/2024-01-30-llava-next/},
    author={Liu, Haotian and Li, Chunyuan and Li, Yuheng and Li, Bo and Zhang, Yuanhan and Shen, Sheng and Lee, Yong Jae},
    month={January},
    year={2024}
}

@article{videochatgpt,
  title={Video-chatgpt: Towards detailed video understanding via large vision and language models},
  author={Maaz, Muhammad and Rasheed, Hanoona and Khan, Salman and Khan, Fahad Shahbaz},
  journal={arXiv preprint arXiv:2306.05424},
  year={2023}
}

@article{videochat,
  title={Videochat: Chat-centric video understanding},
  author={Li, KunChang and He, Yinan and Wang, Yi and Li, Yizhuo and Wang, Wenhai and Luo, Ping and Wang, Yali and Wang, Limin and Qiao, Yu},
  journal={arXiv preprint arXiv:2305.06355},
  year={2023}
}

@article{videollava,
  title={Video-llava: Learning united visual representation by alignment before projection},
  author={Lin, Bin and Zhu, Bin and Ye, Yang and Ning, Munan and Jin, Peng and Yuan, Li},
  journal={arXiv preprint arXiv:2311.10122},
  year={2023}
}

@article{longva,
  title={Long context transfer from language to vision},
  author={Zhang, Peiyuan and Zhang, Kaichen and Li, Bo and Zeng, Guangtao and Yang, Jingkang and Zhang, Yuanhan and Wang, Ziyue and Tan, Haoran and Li, Chunyuan and Liu, Ziwei},
  journal={arXiv preprint arXiv:2406.16852},
  year={2024}
}

@article{mmvid,
  title={Mm-vid: Advancing video understanding with gpt-4v (ision)},
  author={Lin, Kevin and Ahmed, Faisal and Li, Linjie and Lin, Chung-Ching and Azarnasab, Ehsan and Yang, Zhengyuan and Wang, Jianfeng and Liang, Lin and Liu, Zicheng and Lu, Yumao and others},
  journal={arXiv preprint arXiv:2310.19773},
  year={2023}
}

@inproceedings{videoagent,
  title={Videoagent: A memory-augmented multimodal agent for video understanding},
  author={Fan, Yue and Ma, Xiaojian and Wu, Rujie and Du, Yuntao and Li, Jiaqi and Gao, Zhi and Li, Qing},
  booktitle={European Conference on Computer Vision},
  pages={75--92},
  year={2025},
  organization={Springer}
}

@misc{longllava,
author = { {Yin Song and Chen Wu and Eden Duthie} },
title = { {aws-prototyping/long-llava-qwen2-7b} },
year = 2024,
url = { https://huggingface.co/aws-prototyping/long-llava-qwen2-7b },
publisher = { Hugging Face }
}

@inproceedings{clip,
  title={Learning transferable visual models from natural language supervision},
  author={Radford, Alec and Kim, Jong Wook and Hallacy, Chris and Ramesh, Aditya and Goh, Gabriel and Agarwal, Sandhini and Sastry, Girish and Askell, Amanda and Mishkin, Pamela and Clark, Jack and others},
  booktitle={International conference on machine learning},
  pages={8748--8763},
  year={2021},
  organization={PMLR}
}

@article{faiss,
  title={Billion-scale similarity search with GPUs},
  author={Johnson, Jeff and Douze, Matthijs and J{\'e}gou, Herv{\'e}},
  journal={IEEE Transactions on Big Data},
  volume={7},
  number={3},
  pages={535--547},
  year={2019},
  publisher={IEEE}
}

@article{videomme,
  title={Video-MME: The First-Ever Comprehensive Evaluation Benchmark of Multi-modal LLMs in Video Analysis},
  author={Fu, Chaoyou and Dai, Yuhan and Luo, Yondong and Li, Lei and Ren, Shuhuai and Zhang, Renrui and Wang, Zihan and Zhou, Chenyu and Shen, Yunhang and Zhang, Mengdan and others},
  journal={arXiv preprint arXiv:2405.21075},
  year={2024}
}

@inproceedings{whisper,
  title={Robust speech recognition via large-scale weak supervision},
  author={Radford, Alec and Kim, Jong Wook and Xu, Tao and Brockman, Greg and McLeavey, Christine and Sutskever, Ilya},
  booktitle={International conference on machine learning},
  pages={28492--28518},
  year={2023},
  organization={PMLR}
}

@article{contriever,
  title={Unsupervised dense information retrieval with contrastive learning},
  author={Izacard, Gautier and Caron, Mathilde and Hosseini, Lucas and Riedel, Sebastian and Bojanowski, Piotr and Joulin, Armand and Grave, Edouard},
  journal={arXiv preprint arXiv:2112.09118},
  year={2021}
}

@inproceedings{ape,
  title={Aligning and prompting everything all at once for universal visual perception},
  author={Shen, Yunhang and Fu, Chaoyou and Chen, Peixian and Zhang, Mengdan and Li, Ke and Sun, Xing and Wu, Yunsheng and Lin, Shaohui and Ji, Rongrong},
  booktitle={Proceedings of the IEEE/CVF Conference on Computer Vision and Pattern Recognition},
  pages={13193--13203},
  year={2024}
}

@article{qwen2vl,
  title={Qwen2-VL: Enhancing Vision-Language Model's Perception of the World at Any Resolution},
  author={Wang, Peng and Bai, Shuai and Tan, Sinan and Wang, Shijie and Fan, Zhihao and Bai, Jinze and Chen, Keqin and Liu, Xuejing and Wang, Jialin and Ge, Wenbin and others},
  journal={arXiv preprint arXiv:2409.12191},
  year={2024}
}

@misc{llavavideo, 
    title={Video Instruction Tuning With Synthetic Data}, author={Yuanhan Zhang and Jinming Wu and Wei Li and Bo Li and Zejun Ma and Ziwei Liu and Chunyuan Li}, 
    year={2024}, 
    eprint={2410.02713}, 
    archivePrefix={arXiv}, 
    primaryClass={cs.CV}, 
    url={https://arxiv.org/abs/2410.02713}, }

@article{chatunivi,
  title={Chat-UniVi: Unified Visual Representation Empowers Large Language Models with Image and Video Understanding}, 
  author={Peng Jin and Ryuichi Takanobu and Caiwan Zhang and Xiaochun Cao and Li Yuan},
  journal={arXiv preprint arXiv:2311.08046},
  year={2023}
}

@article{mlvu,
  title={MLVU: A Comprehensive Benchmark for Multi-Task Long Video Understanding},
  author={Zhou, Junjie and Shu, Yan and Zhao, Bo and Wu, Boya and Xiao, Shitao and Yang, Xi and Xiong, Yongping and Zhang, Bo and Huang, Tiejun and Liu, Zheng},
  journal={arXiv preprint arXiv:2406.04264},
  year={2024}
}

@article{gemini,
  title={Gemini 1.5: Unlocking multimodal understanding across millions of tokens of context},
  author={Reid, Machel and Savinov, Nikolay and Teplyashin, Denis and Lepikhin, Dmitry and Lillicrap, Timothy and Alayrac, Jean-baptiste and Soricut, Radu and Lazaridou, Angeliki and Firat, Orhan and Schrittwieser, Julian and others},
  journal={arXiv preprint arXiv:2403.05530},
  year={2024}
}

@misc{gpt4o,
  author    = {OpenAI},
  title     = {GPT-4o System Card},
  year      = {2024},
  howpublished = {\url{https://openai.com/index/gpt-4o-system-card/}},
}

@article{intp,
  title={Interpolating Video-LLMs: Toward Longer-sequence LMMs in a Training-free Manner},
  author={Shang, Yuzhang and Xu, Bingxin and Kang, Weitai and Cai, Mu and Li, Yuheng and Wen, Zehao and Dong, Zhen and Keutzer, Kurt and Lee, Yong Jae and Yan, Yan},
  journal={arXiv preprint arXiv:2409.12963},
  year={2024}
}

@article{rag,
  title={Retrieval-augmented generation for knowledge-intensive nlp tasks},
  author={Lewis, Patrick and Perez, Ethan and Piktus, Aleksandra and Petroni, Fabio and Karpukhin, Vladimir and Goyal, Naman and K{\"u}ttler, Heinrich and Lewis, Mike and Yih, Wen-tau and Rockt{\"a}schel, Tim and others},
  journal={Advances in Neural Information Processing Systems},
  volume={33},
  pages={9459--9474},
  year={2020}
}

@article{irag,
  title={iRAG: An Incremental Retrieval Augmented Generation System for Videos},
  author={Arefeen, Md Adnan and Debnath, Biplob and Uddin, Md Yusuf Sarwar and Chakradhar, Srimat},
  journal={arXiv preprint arXiv:2404.12309},
  year={2024}
}

@article{omagent,
  title={Omagent: A multi-modal agent framework for complex video understanding with task divide-and-conquer},
  author={Zhang, Lu and Zhao, Tiancheng and Ying, Heting and Ma, Yibo and Lee, Kyusong},
  journal={arXiv preprint arXiv:2406.16620},
  year={2024}
}

@inproceedings{gupta2023visual,
  title={Visual programming: Compositional visual reasoning without training},
  author={Gupta, Tanmay and Kembhavi, Aniruddha},
  booktitle={Proceedings of the IEEE/CVF Conference on Computer Vision and Pattern Recognition},
  pages={14953--14962},
  year={2023}
}

@inproceedings{suris2023vipergpt,
  title={Vipergpt: Visual inference via python execution for reasoning},
  author={Sur{\'\i}s, D{\'\i}dac and Menon, Sachit and Vondrick, Carl},
  booktitle={Proceedings of the IEEE/CVF International Conference on Computer Vision},
  pages={11888--11898},
  year={2023}
}

@article{videollama,
  title={Video-llama: An instruction-tuned audio-visual language model for video understanding},
  author={Zhang, Hang and Li, Xin and Bing, Lidong},
  journal={arXiv preprint arXiv:2306.02858},
  year={2023}
}

@article{vnbench,
  title={Needle In A Video Haystack: A Scalable  Synthetic Framework for Benchmarking Video MLLMs},
  author={Zhao, Zijia and Lu, Haoyu and Huo, Yuqi and Du, Yifan and Yue, Tongtian and Guo, Longteng and Wang, Bingning and Chen, Weipeng and Liu, Jing},
  journal={arXiv preprint},
  year={2024}
}

@inproceedings{stllm,
  title={St-llm: Large language models are effective temporal learners},
  author={Liu, Ruyang and Li, Chen and Tang, Haoran and Ge, Yixiao and Shan, Ying and Li, Ge},
  booktitle={European Conference on Computer Vision},
  pages={1--18},
  year={2025},
  organization={Springer}
}

@inproceedings{vila,
  title={Vila: On pre-training for visual language models},
  author={Lin, Ji and Yin, Hongxu and Ping, Wei and Molchanov, Pavlo and Shoeybi, Mohammad and Han, Song},
  booktitle={Proceedings of the IEEE/CVF Conference on Computer Vision and Pattern Recognition},
  pages={26689--26699},
  year={2024}
}

@inproceedings{internvl,
  title={Internvl: Scaling up vision foundation models and aligning for generic visual-linguistic tasks},
  author={Chen, Zhe and Wu, Jiannan and Wang, Wenhai and Su, Weijie and Chen, Guo and Xing, Sen and Zhong, Muyan and Zhang, Qinglong and Zhu, Xizhou and Lu, Lewei and others},
  booktitle={Proceedings of the IEEE/CVF Conference on Computer Vision and Pattern Recognition},
  pages={24185--24198},
  year={2024}
}

@article{drvideo,
  title={DrVideo: Document Retrieval Based Long Video Understanding},
  author={Ma, Ziyu and Gou, Chenhui and Shi, Hengcan and Sun, Bin and Li, Shutao and Rezatofighi, Hamid and Cai, Jianfei},
  journal={arXiv preprint arXiv:2406.12846},
  year={2024}
}

@article{wang2024longllava,
  title={LongLLaVA: Scaling Multi-modal LLMs to 1000 Images Efficiently via Hybrid Architecture},
  author={Wang, Xidong and Song, Dingjie and Chen, Shunian and Zhang, Chen and Wang, Benyou},
  journal={arXiv preprint arXiv:2409.02889},
  year={2024}
}

@article{chen2024sharegpt4video,
  title={ShareGPT4Video: Improving Video Understanding and Generation with Better Captions},
  author={Chen, Lin and Wei, Xilin and Li, Jinsong and Dong, Xiaoyi and Zhang, Pan and Zang, Yuhang and Chen, Zehui and Duan, Haodong and Lin, Bin and Tang, Zhenyu and others},
  journal={arXiv preprint arXiv:2406.04325},
  year={2024}
}

@article{zhang2024beyond,
  title={Beyond LLaVA-HD: Diving into High-Resolution Large Multimodal Models},
  author={Zhang, Yi-Fan and Wen, Qingsong and Fu, Chaoyou and Wang, Xue and Zhang, Zhang and Wang, Liang and Jin, Rong},
  journal={arXiv preprint arXiv:2406.08487},
  year={2024}
}

@article{kangaroogroup,
	title={Kangaroo: A Powerful Video-Language Model Supporting Long-context Video Input},
	author={Liu, Jiajun and Wang, Yibing and Ma, Hanghang and Wu, Xiaoping and Ma, Xiaoqi and Wei, xiaoming and Jiao, Jianbin and Wu, Enhua and Hu, Jie},
	journal={arXiv preprint arXiv:2408.15542},
	year={2024}
}

@article{liu2024oryx,
  title={Oryx MLLM: On-Demand Spatial-Temporal Understanding at Arbitrary Resolution},
  author={Liu, Zuyan and Dong, Yuhao and Liu, Ziwei and Hu, Winston and Lu, Jiwen and Rao, Yongming},
  journal={arXiv preprint arXiv:2409.12961},
  year={2024}
  }

@article{zhang2023simple,
  title={A simple llm framework for long-range video question-answering},
  author={Zhang, Ce and Lu, Taixi and Islam, Md Mohaiminul and Wang, Ziyang and Yu, Shoubin and Bansal, Mohit and Bertasius, Gedas},
  journal={arXiv preprint arXiv:2312.17235},
  year={2023}
}

@article{wang2024videoagent,
  title={Videoagent: Long-form video understanding with large language model as agent},
  author={Wang, Xiaohan and Zhang, Yuhui and Zohar, Orr and Yeung-Levy, Serena},
  journal={arXiv preprint arXiv:2403.10517},
  year={2024}
}

@article{lvb,
  title={Longvideobench: A benchmark for long-context interleaved video-language understanding},
  author={Wu, Haoning and Li, Dongxu and Chen, Bei and Li, Junnan},
  journal={arXiv preprint arXiv:2407.15754},
  year={2024}
}

@article{pllava,
  title={Pllava: Parameter-free llava extension from images to videos for video dense captioning},
  author={Xu, Lin and Zhao, Yilin and Zhou, Daquan and Lin, Zhijie and Ng, See Kiong and Feng, Jiashi},
  journal={arXiv preprint arXiv:2404.16994},
  year={2024}
}

@article{videoccam,
  title={Video-ccam: Enhancing video-language understanding with causal cross-attention masks for short and long videos},
  author={Fei, Jiajun and Li, Dian and Deng, Zhidong and Wang, Zekun and Liu, Gang and Wang, Hui},
  journal={arXiv preprint arXiv:2408.14023},
  year={2024}
}

@article{videoxl,
  title={Video-XL: Extra-Long Vision Language Model for Hour-Scale Video Understanding},
  author={Shu, Yan and Zhang, Peitian and Liu, Zheng and Qin, Minghao and Zhou, Junjie and Huang, Tiejun and Zhao, Bo},
  journal={arXiv preprint arXiv:2409.14485},
  year={2024}
}

@article{oryx,
title={Oryx MLLM: On-Demand Spatial-Temporal Understanding at Arbitrary Resolution},
author={Liu, Zuyan and Dong, Yuhao and Liu, Ziwei and Hu, Winston and Lu, Jiwen and Rao, Yongming},
journal={arXiv preprint arXiv:2409.12961},
year={2024}
}

@article{aria,
  title={Aria: An Open Multimodal Native Mixture-of-Experts Model}, 
  author={Dongxu Li and Yudong Liu and Haoning Wu and Yue Wang and Zhiqi Shen and Bowen Qu and Xinyao Niu and Guoyin Wang and Bei Chen and Junnan Li},
  year={2024},
  journal={arXiv preprint arXiv:2410.05993},
}

@inproceedings{li2025llama,
  title={Llama-vid: An image is worth 2 tokens in large language models},
  author={Li, Yanwei and Wang, Chengyao and Jia, Jiaya},
  booktitle={European Conference on Computer Vision},
  pages={323--340},
  year={2025},
  organization={Springer}
}

@inproceedings{gradcam,
  title={Grad-cam: Visual explanations from deep networks via gradient-based localization},
  author={Selvaraju, Ramprasaath R and Cogswell, Michael and Das, Abhishek and Vedantam, Ramakrishna and Parikh, Devi and Batra, Dhruv},
  booktitle={Proceedings of the IEEE international conference on computer vision},
  pages={618--626},
  year={2017}
}

@article{tsne,
  title={Visualizing data using t-SNE.},
  author={Van der Maaten, Laurens and Hinton, Geoffrey},
  journal={Journal of machine learning research},
  volume={9},
  number={11},
  year={2008}
}

@inproceedings{zong2024long,
  title={Long-context vision large language models: Empirical insights and a baseline},
  author={Zong, Yongshuo and Elezi, Ismail and Yang, Yongxin and Deng, Jiankang and Hospedales, Timothy},
  booktitle={Workshop on Long Context Foundation Models},
  year={2024}
}

@article{videorag,
  title={VideoRAG: Retrieval-Augmented Generation with Extreme Long-Context Videos},
  author={Ren, Xubin and Xu, Lingrui and Xia, Long and Wang, Shuaiqiang and Yin, Dawei and Huang, Chao},
  journal={arXiv preprint arXiv:2502.01549},
  year={2025}
}

@article{jeong2025videorag,
  title={Videorag: Retrieval-augmented generation over video corpus},
  author={Jeong, Soyeong and Kim, Kangsan and Baek, Jinheon and Hwang, Sung Ju},
  journal={arXiv preprint arXiv:2501.05874},
  year={2025}
}

@article{vita15,
  title={Vita-1.5: Towards gpt-4o level real-time vision and speech interaction},
  author={Fu, Chaoyou and Lin, Haojia and Wang, Xiong and Zhang, Yi-Fan and Shen, Yunhang and Liu, Xiaoyu and Cao, Haoyu and Long, Zuwei and Gao, Heting and Li, Ke and others},
  journal={arXiv preprint arXiv:2501.01957},
  year={2025}
}

\newpage
\setlength{\parindent}{0pt}

\begin{center}
    \LARGE\bfseries Supplemental Material
\end{center}

\appendix
\section{Decouple Query}

\begin{figure*}[h!]
  \centering 
  \includegraphics[width=0.9\linewidth]{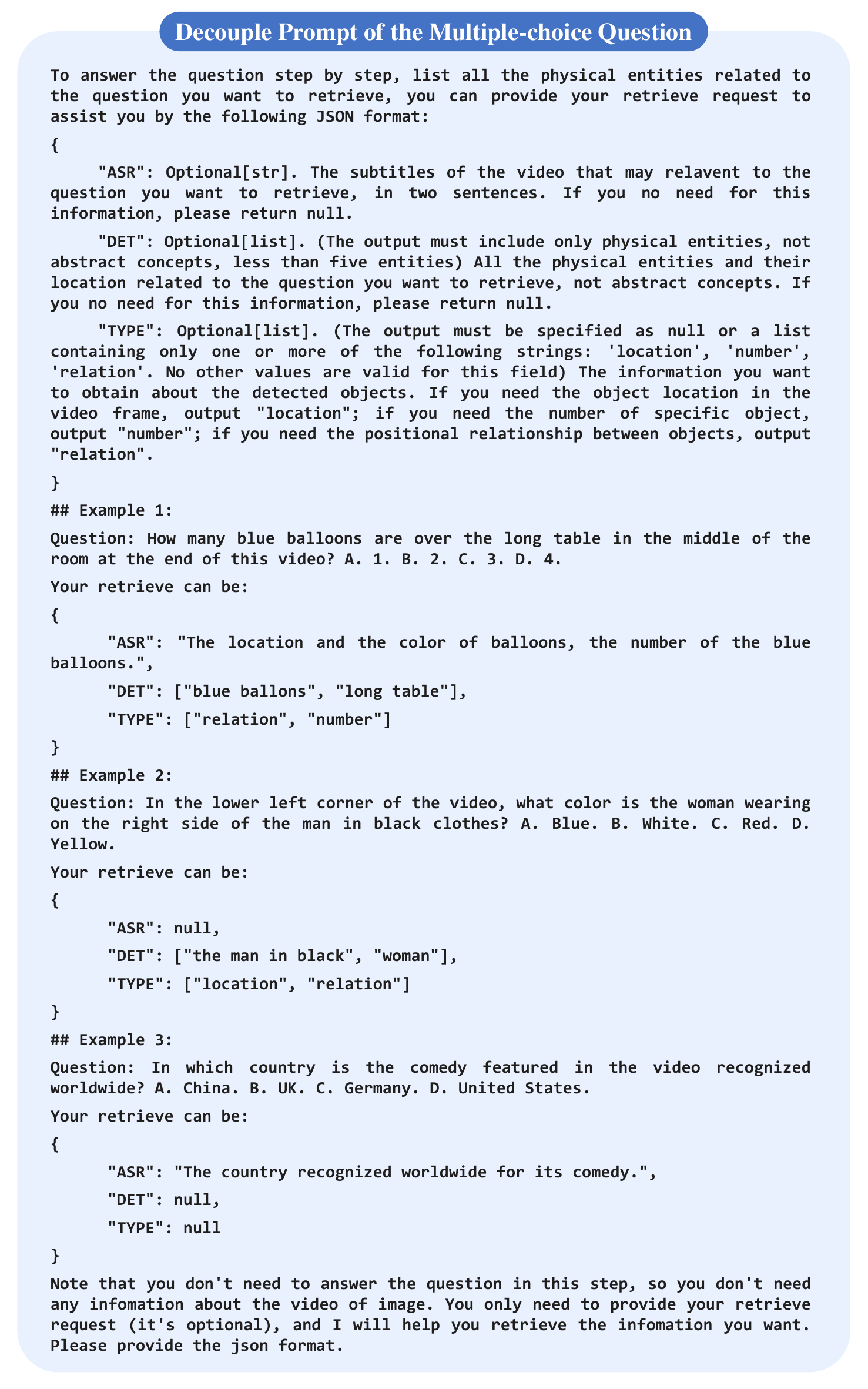}
  \caption{Decouple prompt of the multiple-choice question for LVLMs.}
  \label{fig_decouple} 
\end{figure*}

In the initial phase of the proposed Video-RAG, we employ a decouple prompt, denoted as $\bm{\mathrm{P}}$, to guide the LVLM in generating retrieval requests. In this section, we present one example of a prompt designed for multiple-choice questions, as illustrated in Figure \ref{fig_decouple}.

\section{Sub-set of Video-MME}
\label{subset}

As outlined in the implementation details, we randomly sampled a subset of the Video-MME \cite{videomme} dataset to evaluate a computationally resource-intensive, agent-based method with long-context LVLMs. Specifically, we selected 10\% of the full dataset, comprising 30 short, 30 medium-length, and 30 long videos. Each video contains three multiple-choice questions. Importantly, we ensured that the performance ranking of the methods on the subset mirrored that of the full dataset. As shown in Tables \ref{tab_mini} and \ref{tab_mini_2}, we evaluated four distinct 7B models Chat-Univi-v1.5 \cite{chatunivi}, LLaVA-NeXT-Video \cite{llavanextvideo}, LongVA \cite{longva}, and Long-LLaVA \cite{longllava} using a frame sampling rate of 16 for both the subset and the full set. Our results indicate that the performance rankings remained consistent across both evaluations.

\begin{table}[h]
\centering 
\caption{ Performance of Video-MME sub-set.}
\setlength{\tabcolsep}{0.8mm} 
\renewcommand\arraystretch{1.0} 
\begin{tabular}{l|ccc|c}
\toprule
\textbf{Method} & \textbf{Short} & \textbf{Medium}  & \textbf{Long} & \textbf{Overall}   \\ \midrule
Chat-Univi-v1.5 \cite{chatunivi}  & 50.0 & 33.3 & 17.8 & 33.7    \\ 
LLaVA-NeXT-Video \cite{llavanextvideo}  & 54.4 & 33.3 & 23.3 & 37.0    \\ 
LongVA \cite{longva} & 56.7 & 50.0 & 38.9 & 48.5    \\ 
Long-LLaVA \cite{longllava} & 58.9 & 52.2 & 40.0 & 50.4    \\ 
\bottomrule
\end{tabular}

\label{tab_mini}
\end{table}

\begin{table}[h]
\centering 
\caption{ Performance of Video-MME full-set.}
\setlength{\tabcolsep}{0.8mm} 
\renewcommand\arraystretch{1.0} 
\begin{tabular}{l|ccc|c}
\toprule
\textbf{Method} & \textbf{Short} & \textbf{Medium}  & \textbf{Long} & \textbf{Overall}   \\ \midrule
Chat-Univi-v1.5 \cite{chatunivi} & 45.7 & 39.0 & 35.7 & 40.1    \\ 
LLaVA-NeXT-Video \cite{llavanextvideo} & 51.1 & 41.8 & 36.8 & 43.2    \\ 
LongVA \cite{longva} & 60.8 & 45.2 & 41.4 & 49.1    \\ 
Long-LLaVA \cite{longllava} & 59.3 & 49.3 & 44.4 & 51.0    \\ 
\bottomrule
\end{tabular}

\label{tab_mini_2}
\end{table}

\section{Results on Video-MME Sub-Set}

\begin{figure}[t!]
  \centering 
  \includegraphics[width=0.6\linewidth]{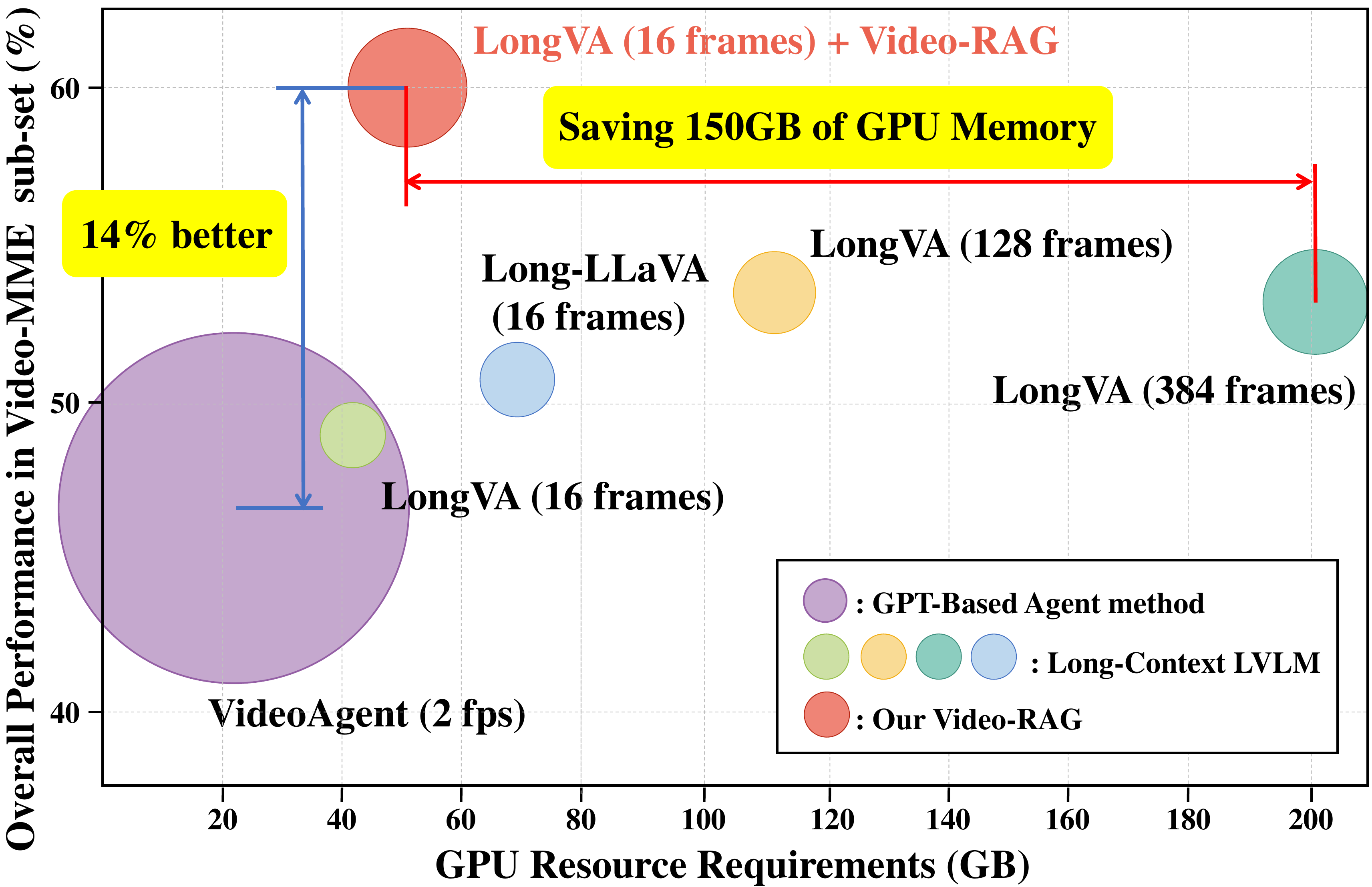}
  \caption{The comparison of our Video-RAG with two common approaches. The size of the bubbles represents the total time consumed for completing inference on the Video-MME \cite{videomme} sub-set.}
  \label{fig_cmp} 
\end{figure}

We examine Video-RAG against two representative methods in terms of inference time, GPU resource requirements, and overall performance. Given that GPT-based Agent methods are resource-intensive, we randomly sampled a sub-set of the Video-MME \cite{videomme} for evaluation, as described in Section \ref{subset}.
As demonstrated in Figure \ref{fig_cmp}, VideoAgent \cite{videoagent}, a typically GPT-based Agent method, requires significant time to process video and deliver suboptimal performance. Meanwhile, LongVA \cite{longva}, a representative long-context LVLM, shows limited improvement from increasing the frame rate and even experiences performance degradation. Integrating our Video-RAG into the 16-frame LongVA results in substantial performance improvements while reducing GPU resource consumption. Specifically, with only increasing 8GB GPU memory compared to the base (16-frames LongVA), we achieve 11.5\% overall performance improvement, while outperforming another long-context LVLM Long-LLaVA-7B \cite{longllava} in 16-frames setting by 9.6\% with less GPU memory requirements and compatible total inference time. These results demonstrated that our Video-RAG is lightweight with lower computing overhead than the other typical methods. Moreover, we provide detailed time consuming to construct three types of databases (which can be built in parallel) and inference per query, as shown in Table \ref{time}.

\begin{table}[h]
\vspace{-0.9em}
\centering
\caption{\footnotesize{Overall performance, databases construct and average inference time (include building databases) per query (\#Time) in Video-MME-mini.}}
\footnotesize
{
\setlength{\tabcolsep}{1.5pt}
{
\begin{tabular}{l|cccc|cc|cc}
\toprule
\multirow{2}{*}{Model} & \multirow{2}{*}{ASR} & \multirow{2}{*}{OCR} & \multirow{2}{*}{DET} & \multirow{2}{*}{Total Time} & \multicolumn{2}{c|}{w/o subs} & \multicolumn{2}{c}{w/ Video-RAG} \\
  &  &  &  & &  \#Time & Overall & \#Time & Overall  \\ 
\midrule
\specialrule{0em}{1pt}{0pt} \footnotesize{VideoAgent} & - & - & - & - &14min & 47.7 & - & -\\
\specialrule{0em}{0pt}{0pt} \footnotesize{LongVA-16fs} & 21min & 2min & 3min & max(21, 2, 3)=21min & \color{cblue}1s & 48.5 & \color{cblue}{1s} \color{black}+ \color{cgreen}5s & 60.0  \\
\specialrule{0em}{0pt}{0pt} \footnotesize{LongVA-128fs} & 21min & 16min & 16min & max(21, 16, 16)=21min & \color{cblue}8s & 54.1 & \color{cblue}{8s} \color{black}+ \color{cgreen}5s & 63.3 \\
 \specialrule{0em}{0pt}{0pt} \footnotesize{LongVA-384fs} & 42min & 48min & 24min & max(42, 48, 24)=48min &  \color{cblue}20s & 53.7 & \color{cblue}{20s} \color{black}+ \color{cgreen}11s & 63.6 \\
\bottomrule
\end{tabular}
}
}

\label{time}
\end{table}

\section{Details of Similarity Score Calculation}

In the process of using the RAG system to retrieve auxiliary texts extracted from videos, we define a similarity threshold $t$ to ensure the selection of relevant texts. Specifically, we employ FAISS-based \cite{faiss} similarity to select OCR and ASR texts, while CLIP \cite{clip} similarity is used for keyframe selection. In our implementation, the similarity threshold $t$ is set to 0.3.
As for OCR and ASR selection, For any given list of the retrieve request \( \bm{\mathrm{R}} \) and auxiliary texts \( \bm{\mathrm{A}} \), the Contriever \cite{contriever} framework maps the text to a text embedding as:

\[ \bm{\mathrm{E}}_{a_i} = \texttt{Contriever}(\bm{\mathrm{A}}_{i}), \quad i = 1, 2, \dots, n \]
\[ \bm{\mathrm{E}}_{r_i} = \texttt{Contriever}(\bm{\mathrm{R}}_{i}), \quad i = 1, 2, \dots, n \]

The average embedding of the retrieve request is then computed as:
\[  \bm{\mathrm{E}}_{r} = \frac{1}{n} \sum_{i=1}^{n} \bm{\mathrm{E}}_{r_i}\]

After that, the embedding of the request and the list of auxiliary texts is normalized:
\[ \bm{\mathrm{E}}_{a_i} = \frac{\bm{\mathrm{E}}_{a_i}}{\|\ \bm{\mathrm{E}}_{a_i}\|}, \quad \bm{\mathrm{E}}_{r} = \frac{\bm{\mathrm{E}}_{r}}{\|\ \bm{\mathrm{E}}_{r}\|}\]

The similarity between the query embedding \( \bm{\mathrm{E}}_{r} \) and the document vector \( \bm{\mathrm{E}}_{a} \) is computed using the inner product, the FAISS library is employed to efficiently perform this search and return the indices of the auxiliary texts meeting the criterion:
\[ S(\bm{\mathrm{E}}_{r}, \bm{\mathrm{E}}_{a_i}) = \bm{\mathrm{E}}_{r} \cdot \bm{\mathrm{E}}_{a_i} > t\]

As for object detection, we use CLIP to select the video keyframe. During this process, we first filter the object detection request $\bm{\mathrm{R}}_{det}$ to ensure they correspond to CLIP-sensitive physical entities, avoiding the inclusion of abstract concepts. Specifically, if it is a single word, direct part-of-speech filtering is applied; if it is a compound word, certain rules are followed to check for compliance, such as whether it is an adjective plus a noun, or a noun plus a noun. We use the Spacy library to achieve this. After this, we put the text ``A picture of" before each object detection request.

Then, we extracting embedding from both the video frames $\bm{\mathrm{F}}$ and the detection request $\bm{\mathrm{R}}_{det}$:

\[ \bm{\mathrm{E}}_{\bm{\mathrm{F}}_j} = \texttt{CLIP}(\bm{\mathrm{F}}_{j}), \quad j = 1, 2, \dots, m \]
\[ \bm{\mathrm{E}}_{\bm{\mathrm{R}}_{i}} = \texttt{CLIP}(\bm{\mathrm{R}}_{det_i}), \quad i = 1, 2, \dots, n \]

The similarity between each video frame and the detection retrieve requests is computed using the dot product between the image and text feature embeddings. For each frame \( \bm{\mathrm{F}}_{j} \), and for each retrieve request \( \bm{\mathrm{E}}_{\bm{\mathrm{R}}_{i}} \), the similarity score is given by:

\[ S_{ij} = \bm{\mathrm{E}}_{\bm{\mathrm{F}}_j} \cdot \bm{\mathrm{E}}_{\bm{\mathrm{R}}_{i}} \]
where \( \cdot \) denotes the dot product. The final similarity score for each frame is the average similarity across all requests:
\[
S_j = \frac{1}{n} \sum_{i=1}^{n} S_{ij}
\]

This computes the mean similarity for each frame across all text descriptions, resulting in a similarity vector \( \mathbf{S} = [S_1, S_2, \dots, S_m] \). The similarity scores are adjusted by a scaling factor \( \alpha \), which is computed based on the number of frames \( m \) and a base frame number \( b \) (which is set to 16 and 4.0, respectively) to adapted different video sampling rate of LVLMs:

\[ \alpha = \beta \times \frac{m}{b} \]
where \( \beta \) is a predefined scaling parameter.

Next, the similarity scores are scaled and normalized to ensure that they sum to 1:
\[
S_j^{\text{norm}} = \frac{\alpha \times S_j}{\sum_{k=1}^{m} S_k}
\]
where \( S_j^{\text{norm}} \) represents the normalized similarity score for frame \( \bm{\mathrm{F}}_{j} \).

The final step is to select the keyframes based on the normalized similarity scores. A threshold \( t \) is applied to the normalized similarities, such that frames with similarity scores above the threshold are selected as keyframes:
\[ \text{Keyframe:} \quad \bm{\mathrm{F}}_{j} \quad \text{if} \quad S_j^{\text{norm}} > t\]
Thus, the set of selected keyframes is given by:
\[
\bm{\mathrm{F}}_{key} = \{\bm{\mathrm{F}}_{j} \mid S_j^{\text{norm}} > t, \, j = 1, 2, \dots, m\}
\]

\section{More Ablation Studies}

\begin{figure}[h]
  \centering 
  \includegraphics[width=0.6\linewidth]{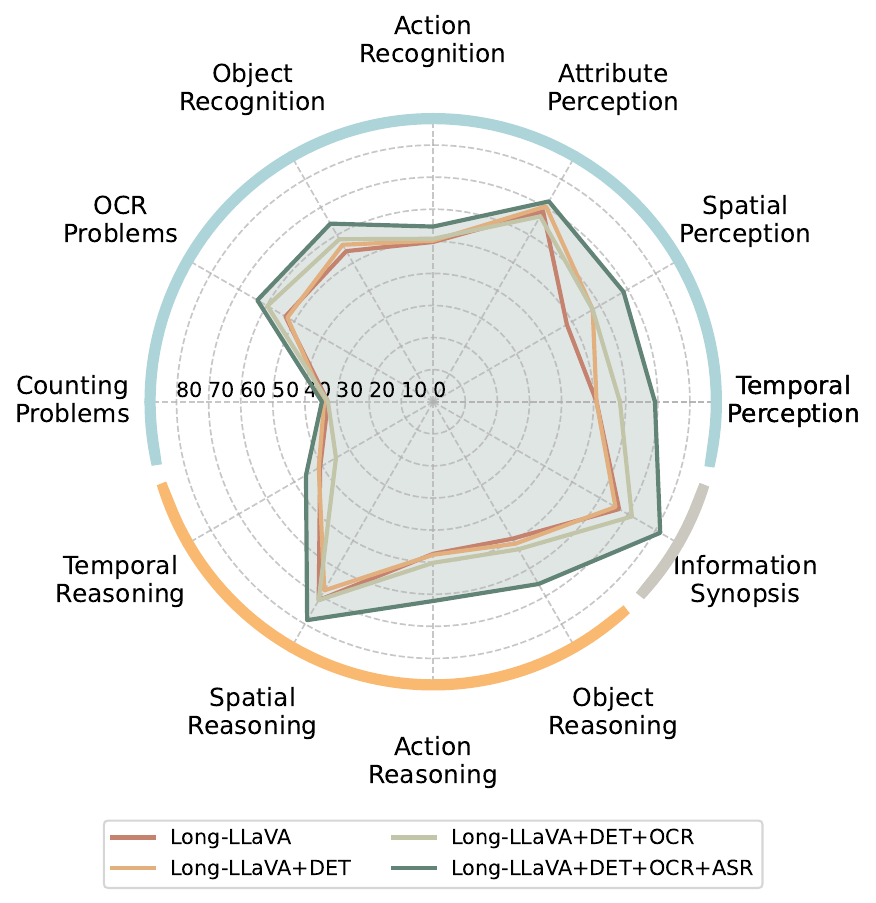}
  \caption{Performance on 12 sub-tasks in Video-MME \cite{videomme} benchmark after applying different components in Long-LLaVA.}
  \label{fig_radar} 
\end{figure}

\noindent \textbf{Effect of different components of Video-RAG.} We evaluate the performance across sub-tasks within Video-MME \cite{videomme}, as shown in Figure \ref{fig_radar}. The results reveal that object detection auxiliary texts significantly enhance spatial perception and object counting, while OCR auxiliary texts specifically improve performance on text recognition tasks. Additionally, ASR auxiliary texts contribute to a general improvement in inference tasks, underscoring the critical role of audio transcription in video understanding. Given that audio transcription is considerably more time-consuming than character recognition or object detection, these texts should be selected based on the requirements of the application.

Besides studying the inference of different components of Video-RAG in the Video-MME \cite{videomme} benchmark, we also experiment with a different type of video benchmark.
We first evaluate LLaVA-Video in MLVU \cite{mlvu} and LongVideoBench \cite{lvb} in both 7B and 72B scale with the 64-frame setting, results are shown in Table \ref{abl2}. As demonstrated, when all components are combined, we get optimal performance in both datasets, including 7B and 72B scales. Specifically, the performance in MLVU \cite{mlvu} even declined when the RAG system was not implemented.

\begin{table}[h]
\centering 
\caption{Ablation study in MLVU and LongVideoBench.}
\setlength{\tabcolsep}{1.0mm} 
\renewcommand\arraystretch{0.95} 
\begin{tabular}{cccc|cc|cc}
\toprule
\multirow{2}{*}{RAG} & \multirow{2}{*}{DET} & \multirow{2}{*}{OCR} & \multirow{2}{*}{ASR} & \multicolumn{2}{c|}{7B} & \multicolumn{2}{c}{72B} \\
& & & & MLVU & LVB & MLVU & LVB \\
\midrule
 & & & & 70.8 & 56.6 & 73.1 & 61.9 \\
 \checkmark & \checkmark & & & 71.0 & 56.5 & 73.4 & 63.2 \\
 \checkmark & \checkmark & \checkmark & & 71.3 & 56.8 & 73.5 & 63.4 \\
\rowcolor{cyan!10} \checkmark & \checkmark & \checkmark & \checkmark & \textbf{72.4} & \textbf{58.7} & \textbf{73.8} &\textbf{65.4} \\
 & \checkmark & \checkmark & \checkmark & 70.3 & 58.3 & 72.9 & 64.0 \\
\bottomrule
\end{tabular}

\label{abl2}
\end{table}

Then, to better point out the role of DET and OCR, we evaluate Video-RAG in VNBench \cite{vnbench} with Long-LLaVA-7B \cite{longllava}. VNBench is a synthetic benchmark designed to evaluate models’ long-context abilities, covering tasks such as retrieval, ordering, and counting. VNBench randomly inserts stickers or text into the video that has nothing to do with the original content of the video, thus typically challenging the model's needle-in-the-haystack capability. As shown in Table \ref{abl_vnbench}, we find that applying DET and OCR as auxiliary texts can significantly improve the performance in retrieval, ordering, and counting tasks. However, the ASR component will decline the performance due to the subtitles are not ancillary to this particular task. These results demonstrated that our proposed distinct types of auxiliary texts can be selected according to the application needs to meet the requirements better.

\begin{table}[h]
\centering 
\caption{Results on combinations of different auxiliary texts in VNBench \cite{vnbench} with 1-try setting when applying 7B Long-LLaVA \cite{longllava} as LVLM under the 32-frames setting. \textbf{Ret}, \textbf{Ord}, and \textbf{Cnt} represent retrieval, ordering, and counting tasks, respectively.}
\setlength{\tabcolsep}{1.5mm} 
\renewcommand\arraystretch{0.95} 
\begin{tabular}{c|ccc|ccc|c}
\toprule
 \textbf{RAG} & \textbf{DET} & \textbf{OCR} & \textbf{ASR} & \textbf{Ret} & \textbf{Ord} & \textbf{Cnt} & \textbf{Overall}   \\ \midrule
 &  &   &  & 65.1 & 25.6 & 24.2 & 38.3    \\ 
\checkmark & \checkmark  &  & & 66.9 & 28.4 & 23.8 & 39.7    \\ 
\rowcolor{cyan!10} \checkmark & \checkmark  & \checkmark  &   & 68.2 & 31.3 & 28.9 & 42.8     \\ 
 \checkmark & \checkmark   & \checkmark  & \checkmark  & 66.7 & 31.3 &  29.6 & 42.5     \\
\bottomrule
\end{tabular}

\label{abl_vnbench}
\end{table}

\section{More Qualitative Results}

\begin{figure*}[h!]
  \centering 
  \includegraphics[width=1.0\linewidth]{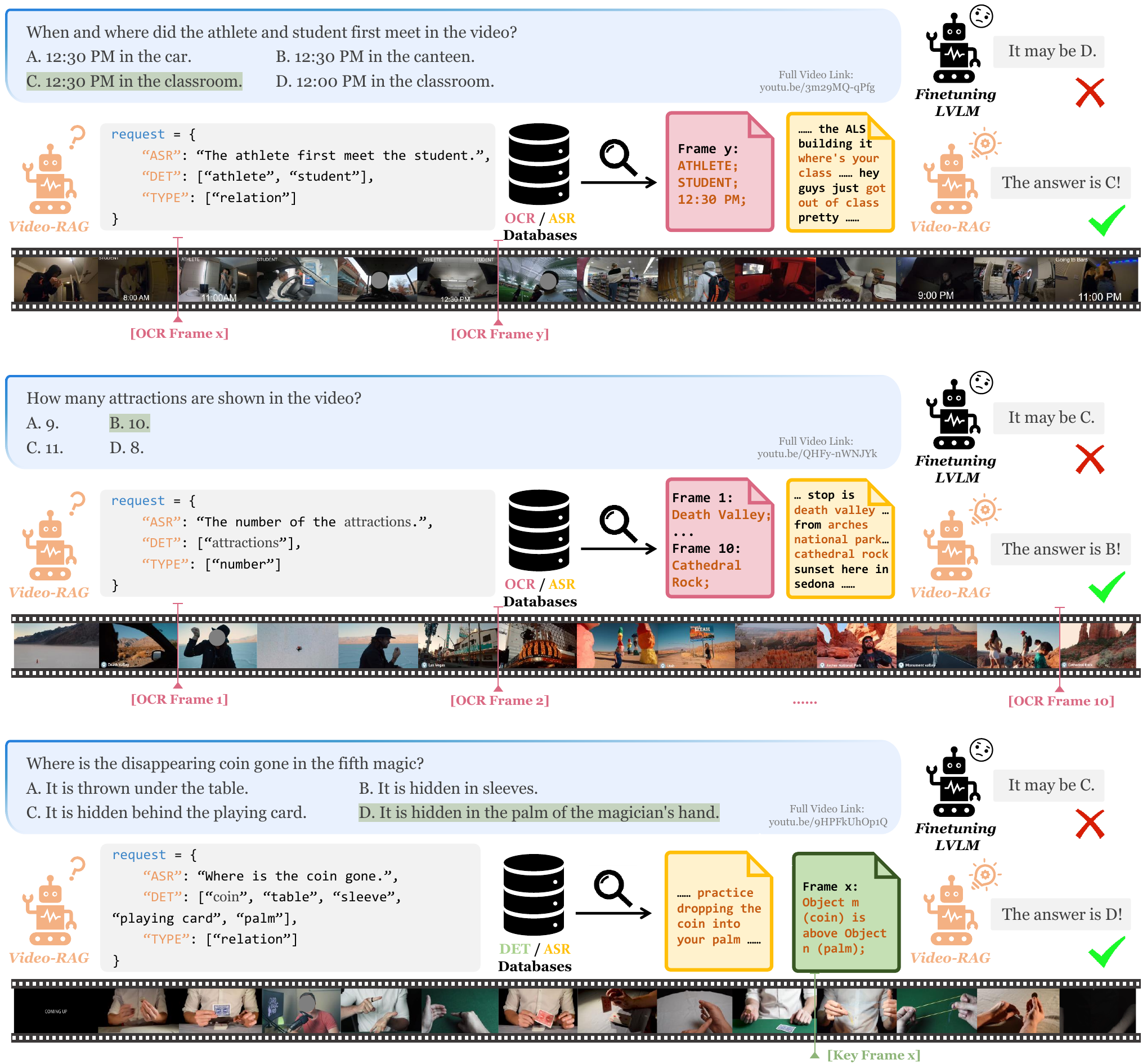}
  \caption{Qualitative results of LLaVA-Vdieo when applying Video-RAG.}
  \label{fig_vis} 
\end{figure*}

In this section, we show more results of LLaVA-Vdieo-7B when applying Video-RAG in different examples in Figure \ref{fig_vis}. The figure highlights several representative cases involving detailed video comprehension from Video-MME \cite{videomme}. As illustrated, augmenting LLaVA-Video with external tools to process and retrieve auxiliary texts from videos significantly enhances its ability to reduce visual hallucinations, thereby enabling more accurate and confident responses to user queries.

\end{document}